\documentclass{article}

\usepackage{iclr2027_conference,times}
\usepackage{amsmath,amssymb,amsfonts}
\usepackage{booktabs}
\usepackage{multirow}
\usepackage{microtype}
\usepackage{graphicx}
\usepackage{subcaption}
\usepackage{xcolor}
\usepackage{colortbl}
\usepackage{enumitem}
\usepackage{pifont}
\usepackage{url}
\usepackage{hyperref}
\hypersetup{
  hidelinks,
  pdftitle={Stable Curves, Unstable Items: Item-Level Scaling Heterogeneity in Video LLMs},
  pdfauthor={Wenzhang Sun, Chunfeng Wang, Xiangchen Yin, Yujia Chen, Hao Li, Kun Zhan}
}

\newcommand{\cmark}{\ding{51}}
\newcommand{\xmark}{\ding{55}}

\newcommand{\eg}{\textit{e.g.}}

\definecolor{oraclecolor}{HTML}{2980B9}

\title{Stable Curves, Unstable Items: Item-Level Scaling Heterogeneity in Video LLMs}

\author{
  Wenzhang Sun \quad
  Chunfeng Wang \quad
  Xiangchen Yin \quad
  Yujia Chen \quad
  Hao Li \quad
  Kun Zhan
}

\iclrfinalcopy

\begin{document}
\maketitle
\fancyhead{}
\renewcommand{\headrulewidth}{0pt}

\begin{abstract}
Aggregate scaling curves suggest that Video LLMs improve smoothly or saturate as visual
budgets grow. We show that this view can conceal large, opposing changes at the item level.
We represent each frozen model--item pair by its response trajectory under controlled visual
budgets and derive matched-grid measures of configuration complementarity, harmful
transitions, and text overwrite. Across five open Video LLMs from three architecture
families, four multiple-choice benchmark splits, open-ended QA and summarization, and
fixed-history dialogue generation, no single budget serves all items. On the four-model
matched MCQA grid, item-level oracle headroom spans $8.8$--$18.9$ accuracy points and
$12.5$--$25.5\%$ of items are correct at a lower budget but wrong at a higher one.
Task-appropriate continuous metrics show the same complementarity beyond multiple choice:
Token-F1 oracle gaps are $2.7$--$3.7$ score points on MLVU generation and $3.8$--$4.8$
points on AVSD current-turn generation, even when mean quality improves with budget.
The effect persists across frame count, spatial resolution, sampling policy,
temporal--spatial allocation, and independently executed raw-video and cached pipelines,
with per-item rates and membership tracking protocol choices. A controlled sampling
intervention recovers $29.0\%$ of terminal regressions, and a structured frame audit identifies
several recurring evidence pathways. We release per-item trajectories, protocol provenance,
derived annotations, and reproducible analysis code as an auditing artifact. A confidence
cascade matches fixed-$128f$ accuracy while reducing average shared frame cost by $31.7\%$,
illustrating one operational use of the response matrix.
\end{abstract}

\section{Introduction}
\label{sec:introduction}

Increasing the number or resolution of visual inputs is a standard way to improve Video
Large Language Models (Video LLMs)~\citep{li2023videochat,zhang2023videollama,
li2024llavaonevision,zhang2024llavavideo,ye2024mplugowl3,li2025f16,
zhang2025flashvid,shu2025videoxl,lin2026stepaudio,sun2026muse}. Evaluation commonly summarizes this intervention with
one aggregate score per budget. Such curves are often smooth or saturating, encouraging the
interpretation that additional visual evidence is beneficial or merely redundant.

\begin{figure}[t]
  \centering
  \includegraphics[width=\linewidth]{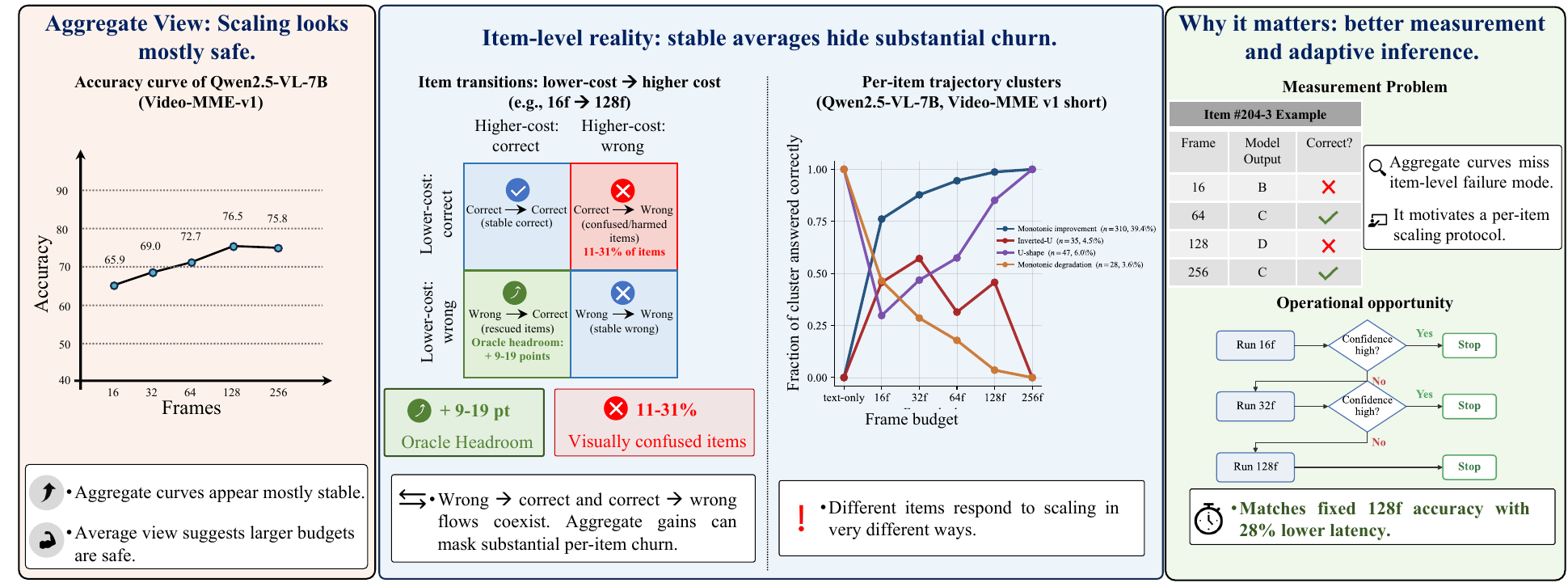}
  \caption{\textbf{A smooth aggregate curve can hide opposing item-level transitions.}
  Qwen2.5-VL-7B on Video-MME v1 short ($n{=}786$). Aggregate accuracy is the net result of
  correct-to-wrong and wrong-to-correct flows; paired trajectories expose the cancellation,
  configuration complementarity, and text-overwrite subsets hidden by the mean.}
  \label{fig:motivation}
\end{figure}

This interpretation does not follow from the mean. At the same budget transition, one item
may flip from wrong to correct while another flips from correct to wrong. These flows can
cancel, making the aggregate curve stable even though the evaluated population changes
substantially. Figure~\ref{fig:motivation} shows this cancellation for Qwen2.5-VL-7B on
Video-MME v1 short. At $128f\!\rightarrow\!256f$, 23 items become wrong and 18 become
correct, yielding only a $-0.7$-point mean change while $5.2\%$ of the benchmark changes
state. We call the resulting bidirectional movement \emph{item-level churn}.

We study this gap through a simple evaluation object: the \emph{paired budget trajectory} of
the same frozen model on the same item under controlled visual interventions. Binary
correctness trajectories yield oracle headroom, visual confusion, text overwrite, and churn.
Continuous task metrics extend the same construction to open-ended generation without
equating lexical score decreases with binary answer flips. Matched grids separate genuine
cross-model differences from unequal architectural frame ceilings, and explicit provenance
prevents outputs from different decoding protocols from being silently pooled.

The expanded evaluation reveals a consistent population-level phenomenon. Across four main
models and four MCQA splits, matched-grid oracle headroom is
$8.8$--$18.9$ points and visual confusion is $12.5$--$25.5\%$. Full-item LLaVA
replications add a third architecture family. On MLVU open-ended QA and summarization and
AVSD fixed-history dialogue, item-level score complementarity remains positive even when the
largest budget improves the average. Resolution, iso-compute allocation, sampling policy,
and independently executed raw/cache grids preserve the phenomenon. Exact rates and item
membership vary with protocol, making per-cell provenance essential to the audit.

Our contributions are:
\begin{itemize}
  \item We formulate Video LLM scaling as paired item-level response trajectories and provide
  matched-grid metrics that expose bidirectional cancellation hidden by aggregate curves.
  \item We establish broad empirical coverage across five frozen open Video LLMs, three
  architecture families, four MCQA splits, open-ended QA, summarization, and controlled
  dialogue generation using metric-appropriate trajectory definitions.
  \item We separate robust population-level structure from protocol-sensitive item labels via
  full sampling, resolution, iso-compute, and raw/cache analyses, and connect a controlled
  sampling intervention to structured failure hypotheses.
  \item We release the per-item response matrix, derived annotations, provenance tags, and
  analysis pipeline, and demonstrate its operational value with a confidence cascade that
  matches fixed-$128f$ accuracy at $31.7\%$ lower shared frame cost.
\end{itemize}

\section{Related Work}
\label{sec:related}

\paragraph{Video LLM scaling.}
Video benchmarks and scaling studies vary frame count, token budget, sampling, or context
length and primarily report aggregate accuracy~\citep{fu2025video,mlvu_2024,
li2024mvbench,liu2024tempcompass,mangalam2023egoschema,zhang2025vps,
li2025afp,huang2025framesampling,zohar2025apollo,li2025f16,wang2023crosssinger}. They establish that more
visual input is not uniformly beneficial at the dataset level. Our focus is different: we ask
which individual items move in each direction, how much opposing movement cancels in the
mean, and whether those trajectories persist across protocols and task formats.

\paragraph{Frame selection, compression, and adaptive inference.}
Frame selectors, clip routers, and token compression methods retain informative evidence
under a fixed budget~\citep{you2025focus,luo2025framevoyager,fan2025air,
hu2025mllmframe,chen2024fastv,chenbeyond}. Confidence cascades and test-time scaling methods allocate
compute from early signals~\citep{wei2022cot,wang2023self,snell2024scaling,
yue2024cascade,chen2024cascadeaware,jitkrittum2025speculative,chen2025fastvlm,
deer_vla_2024,dtoma_2025}. These are methods for choosing or compressing evidence. Our
artifact instead measures the response surface they must navigate. The included cascade is a
compact demonstration of how that response surface supports compute allocation.

\paragraph{Dataset diagnostics and language shortcuts.}
Video-QA audits reveal text-answerable items, language priors, and benchmark artifacts
~\citep{chen2025breakingdown,liu2024mmbench}. Dataset cartography and example-difficulty
work track per-example training dynamics or fixed-compute difficulty
~\citep{swayamdipta2020cartography,baldock2021depth}. We intervene on inference-time
visual budgets for a frozen model. This paired intervention identifies text overwrite and
non-monotonic trajectories that cannot be recovered from a single difficulty score.

\paragraph{Positioning.}
The contribution is therefore an evaluation protocol and reusable response matrix. Relative
to aggregate scaling it exposes direction and cancellation; relative to cartography it changes
the axis from training dynamics to controlled inference budgets; relative to selection methods
it provides the item-level audit target; and relative to text-shortcut analysis it tracks the
entire text-to-video trajectory rather than a single text/video comparison.

\section{Paired Scaling-Trajectory Audit}
\label{sec:setup}
\label{sec:protocol}

\paragraph{Binary trajectories.}
For an item $i$ and an ordered configuration grid $C$, let
$y_{i,c}\in\{0,1\}$ indicate whether a frozen model answers correctly at configuration
$c$. Its trajectory is $\tau_i(C)=(y_{i,c})_{c\in C}$. The best fixed accuracy is
$A_{\mathrm{fixed}}=\max_{c\in C}n^{-1}\sum_i y_{i,c}$ and the item oracle is
$A_{\mathrm{oracle}}=n^{-1}\sum_i\max_{c\in C}y_{i,c}$. Their difference,
\emph{oracle headroom}, measures configuration complementarity. An item is
\emph{visually confused} if $y_{i,c}=1$ at some
lower-cost $c$ and $y_{i,c'}=0$ at some higher-cost $c'$. \emph{Text overwrite} further
requires the lower-cost correct configuration to be text-only. Adjacent churn reports both
$1\!\rightarrow\!0$ and $0\!\rightarrow\!1$ flows rather than their net change.

For any ordered pair $c<c'$, define the rescued and harmed fractions
\begin{equation}
R_{c,c'}=\frac{1}{n}\sum_i(1-y_{i,c})y_{i,c'},\qquad
H_{c,c'}=\frac{1}{n}\sum_i y_{i,c}(1-y_{i,c'}).
\end{equation}
The aggregate change is only their difference,
$A_{c'}-A_c=R_{c,c'}-H_{c,c'}$, whereas pairwise churn is their sum
$R_{c,c'}+H_{c,c'}$. This identity makes the measurement problem explicit: a small net
change does not imply that either directional flow is small.

\paragraph{Continuous trajectories.}
For open-ended generation, correctness is not unambiguous. We replace $y_{i,c}$ by a
task-appropriate score $s_{i,c}\in[0,1]$ and compute the same best-fixed and item-oracle
functionals. We call their difference a \emph{score oracle gap}. A later-budget regression
occurs when a later score is more than $\delta$ below an earlier score; the primary analysis
uses $\delta{=}0.02$ and reports sensitivity at 0.01 and 0.05. We reserve
\emph{visual confusion} and \emph{text overwrite} for binary MCQA and call these continuous
decreases \emph{score regressions}.

\paragraph{Matched grids and provenance.}
Models have different context ceilings. Cross-model MCQA comparisons therefore use the
shared $\{\text{text},16f,32f,64f\}$ grid; LLaVA uses
$\{\text{text},8f,16f,32f\}$ because of its 4K context. Each released cell records model,
split, frame and pixel budgets, sampling method, input size, and raw/cache execution path.
The original Qwen2.5 V1-short anchor trajectory used raw-video decoding, while several later
coverage cells used a 1-fps JPEG cache. We analyze those protocols separately rather than
pooling their item labels.

\begin{table}[t]
\centering
\caption{Evaluation regimes. All reported full cells pass exact-count, no-missing,
no-duplicate, and no-error gates. Additional per-model configuration details are in
Appendix~\ref{app:coverage}.}
\label{tab:evaluation_scope}
\resizebox{\linewidth}{!}{%
\begin{tabular}{llcl}
\toprule
\textbf{Regime} & \textbf{Datasets / items} & \textbf{Models} & \textbf{Primary trajectory signal} \\
\midrule
MCQA & V1 short 786; V1 medium 639; V2 2,516; MLVU 2,170 & 5 & exact option correctness \\
Open generation & MLVU: 201 QA + 217 summaries & 3 & Token-F1; ROUGE-L check \\
Controlled dialogue & AVSD validation: 1,787 turn-10 targets & 2 & Token-F1; ROUGE-L check \\
Setup interventions & V1 short: 786 items / 262 videos & 1 & correctness under paired policies \\
\bottomrule
\end{tabular}%
}
\end{table}

The five frozen open models are Qwen2.5-VL-7B, Qwen3-VL-8B,
InternVL3-8B, InternVL3.5-8B, and LLaVA-NeXT-Video-7B
~\citep{bai2025qwen25vltechnicalreport,bai2026qwen3vl,zhu2025internvl3,
zhu2026internvl35,llava_next_video_2024}. Benchmarks comprise Video-MME v1/v2,
MLVU, and AVSD~\citep{fu2025video,videomme_v2_2026,mlvu_2024,alamri2019audio}.
Unless stated otherwise, inference is greedy and deterministic. Confidence intervals use
5,000 paired item bootstraps that resample item identifiers jointly across configurations.
Paired policy contrasts use McNemar tests; set-overlap nulls preserve the two observed set
sizes exactly; exploratory diagnostic families use permutation tests with BH-FDR correction.

\section{MCQA: Stable Means Hide Unstable Items}
\label{sec:mcqa}
\label{sec:findings}

\paragraph{Cancellation in the anchor grid.}
Qwen2.5-VL-7B accuracy on Video-MME v1 short rises from $65.9\%$ at $16f$ to
$76.5\%$ at $128f$, then changes by only $-0.7$ points at $256f$. This smooth curve is a
net statistic over opposing transitions (Figure~\ref{fig:motivation}). Across adjacent frame
budgets, $5$--$15\%$ of items change correctness. On the time-only raw six-point grid, the
best fixed accuracy is $76.46\%$ and the oracle is $85.24\%$ ($+8.78$ points). Expanding
to the full 11-configuration frame--resolution surface increases the oracle to $87.9\%$
while the best fixed remains $76.5\%$, exposing $+11.5$ points of complementarity.

\begin{figure}[t]
\centering
\includegraphics[width=0.98\linewidth]{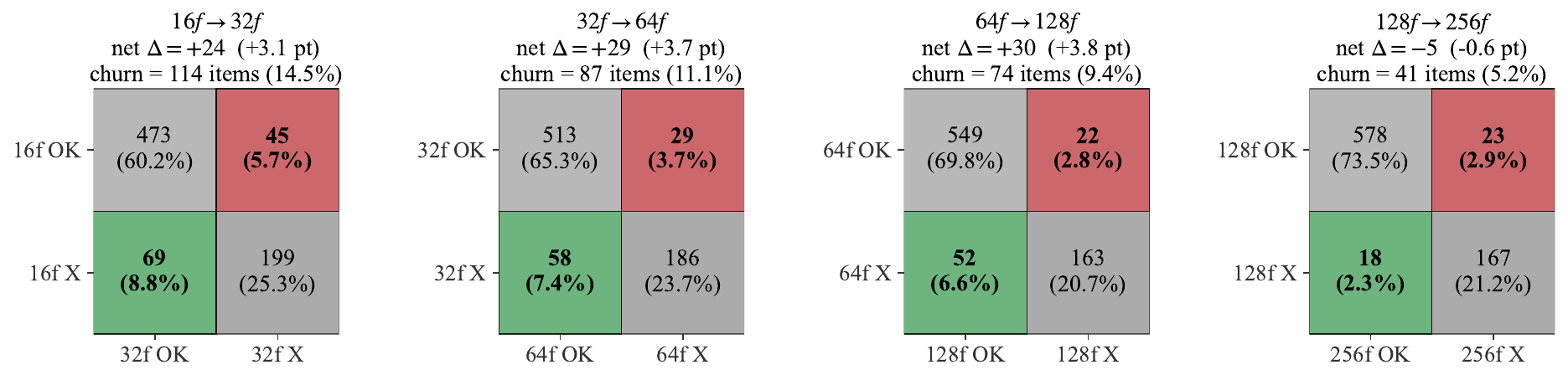}
\caption{\textbf{Directional flows remain substantial even as the aggregate curve
saturates.} Adjacent-budget transition matrices for the Qwen2.5 V1-short raw grid. Green
and red cells are rescued and harmed items; their difference gives the net accuracy change,
while their sum gives churn. At $128f\!\rightarrow\!256f$, a $-0.6$-point net change hides
41 state changes.}
\label{fig:churn_decomposition}
\end{figure}

\begin{figure}[t]
\centering
\includegraphics[width=0.96\linewidth]{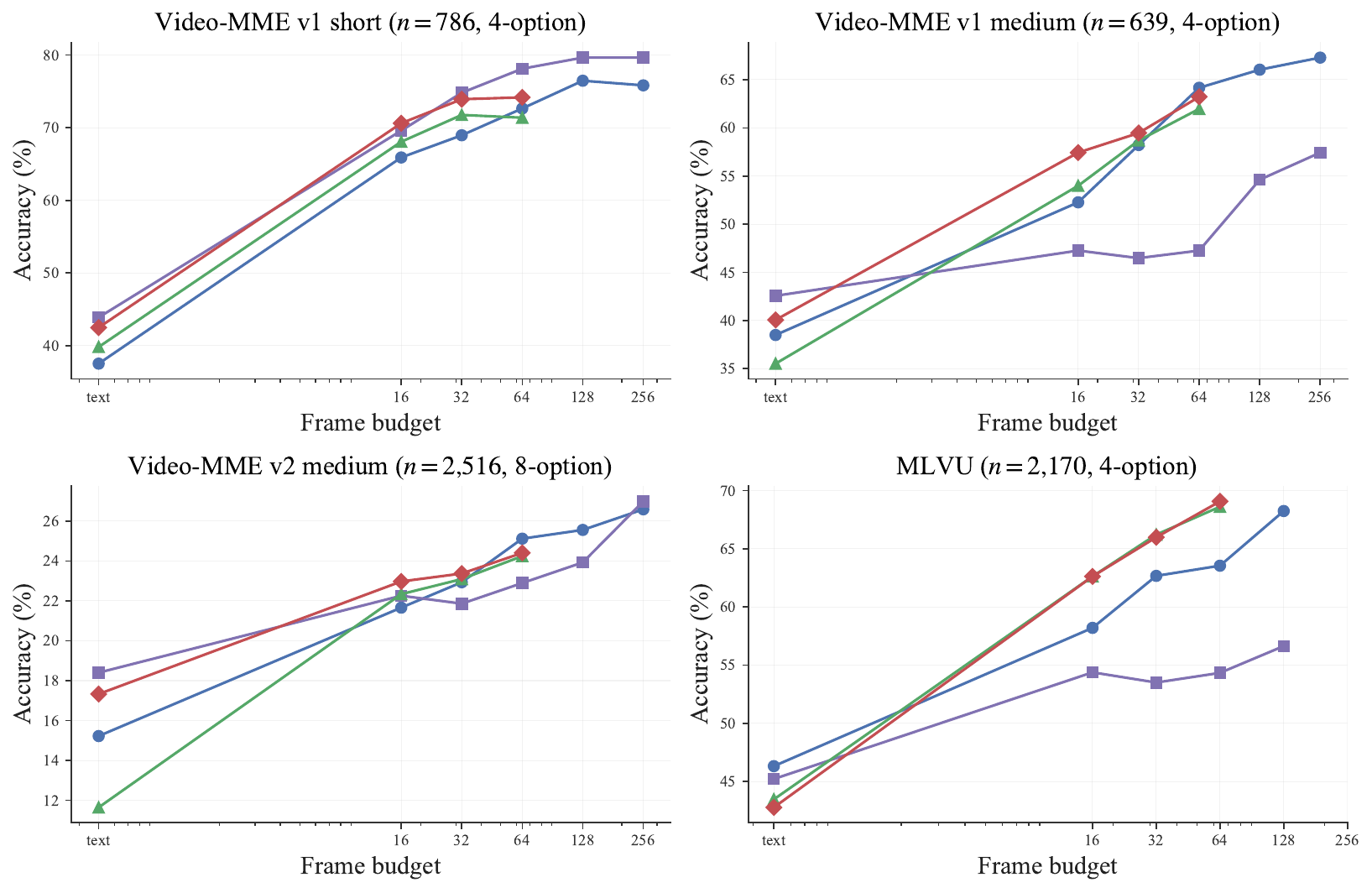}
\caption{\textbf{Aggregate scaling curves across models and benchmark splits.} Curves mostly
improve and saturate, but the paired item-level analysis in Table~\ref{tab:mcqa_grid} reveals
substantial oracle headroom and harmful higher-budget transitions in every matched cell.}
\label{fig:cross_model_curves}
\end{figure}

\paragraph{Cross-model and cross-benchmark replication.}
Table~\ref{tab:mcqa_grid} reports the shared four-point grid for the four main models. Every
cell has $8.8$--$18.9$ points of headroom and $12.5$--$25.5\%$ visual confusion, despite
largely benign aggregate curves. Video-MME v2 changes the answer space from four to eight
options, and MLVU changes benchmark family, yet both retain the same qualitative structure. LLaVA provides a third-family replication under its context-safe
$\{\text{text},8f,16f,32f\}$ grid. On all 786 V1-short items, its best fixed accuracy is
$45.93\%$, oracle accuracy is $57.76\%$, and confusion is $16.16\%$. On all 2,170
MLVU MCQA items, the corresponding values are $48.02\%$, $60.46\%$, and $17.70\%$.
These full-split results extend qualitative replication to a third architecture family under
its supported context grid.

\begin{table}[t]
\centering
\caption{Matched-grid MCQA results. Each entry on the right is oracle headroom in accuracy
points / visual-confusion rate. V1s, V1m, and V2 denote Video-MME v1 short, v1 medium,
and v2 medium.}
\label{tab:mcqa_grid}
\label{tab:crossmodel_summary}
\resizebox{\linewidth}{!}{%
\begin{tabular}{lcccc|cccc}
\toprule
& \multicolumn{4}{c|}{\textbf{Best-fixed accuracy (\%)}} &
\multicolumn{4}{c}{\textbf{Headroom (pts) / confusion (\%)}} \\
\textbf{Model} & V1s & V1m & V2 & MLVU & V1s & V1m & V2 & MLVU \\
\midrule
Qwen2.5-VL-7B  & 72.6 & 64.2 & 25.1 & 63.5 & $+9.8/15.9$ & $+10.6/19.7$ & $+9.6/13.0$ & $+14.5/20.3$ \\
Qwen3-VL-8B    & 78.1 & 47.3 & 22.9 & 54.4 & $+9.5/16.3$ & $+18.9/23.3$ & $+10.6/13.2$ & $+18.6/25.5$ \\
InternVL3-8B   & 71.8 & 62.0 & 24.2 & 68.6 & $+8.8/12.5$ & $+12.8/19.9$ & $+13.6/17.5$ & $+11.6/18.8$ \\
InternVL3.5-8B & 74.2 & 63.2 & 24.4 & 69.1 & $+10.6/14.1$ & $+13.3/20.3$ & $+10.7/13.8$ & $+13.1/20.7$ \\
\bottomrule
\end{tabular}%
}
\end{table}

\paragraph{The heterogeneity has shared and model-specific structure.}
On the Qwen2.5 V1-short raw grid, 151/786 items are visually confused and 39/786 are
correct from text but wrong at $128f$. Among text-correct Counting items, the overwrite rate
reaches $37.1\%$. Non-monotonic trajectory classes account for $19.2\%$ of V1 short and
$18.0\%$ of V1 medium. On the matched V1-short grid, Qwen2.5 and InternVL3 share 33
confused items (Jaccard $17.4\%$), above the $7.5\%$ exact matched-size random expectation
($2.30\times$, $p=2.2\times10^{-6}$). However, 157 of the 190 items in the union occur in
only one model's set. The affected items therefore contain a shared task component alongside
substantial model-specific variation.

\begin{figure}[t]
\centering
\includegraphics[width=0.90\linewidth]{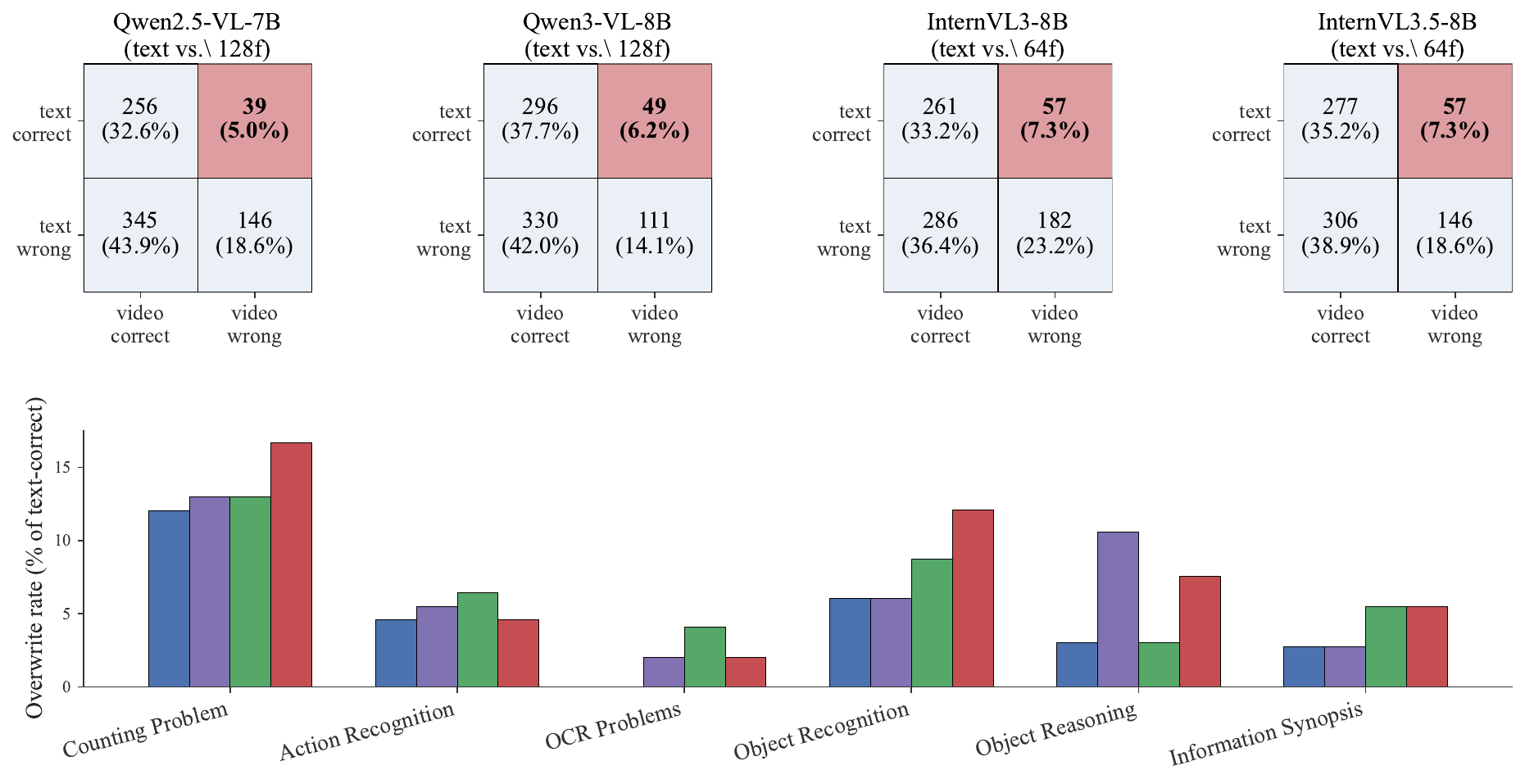}
\caption{\textbf{Text overwrite across four models on Video-MME v1 short.} Top: transitions
from text-only to $128f$ for Qwen models and to the standard-input $64f$ ceiling for InternVL
models. Bottom: overwrite rate among text-correct items for the six largest task types. The
effect is cross-model but strongly task- and model-conditioned.}
\label{fig:overwrite_matrix}
\end{figure}

Text overwrite also replicates across model families (Figure~\ref{fig:overwrite_matrix}).
On V1 short, $5.0$--$7.3\%$ of all items---or $13.2$--$17.9\%$ of items initially
answered correctly from text---become wrong after adding the highest standard-input visual
budget available in the comparison. Counting Problems has the highest conditional overwrite
rate for every model, while the full matrices and cross-model Jaccard show that shared
task-level risk coexists with model-specific overwritten examples.

\section{Beyond Multiple-Choice Accuracy}
\label{sec:generation}

MCQA provides an unambiguous binary contract, but the paired-trajectory construction applies
to any deterministic per-item score. We test two full-split generation settings with Token-F1
as the primary metric and ROUGE-L as a robustness metric. Scores and gaps are displayed on a
0--100 scale; a $+2.7$-point gap means 33.0 best-fixed versus 35.7 oracle, not a $2.7\%$
relative improvement.

\begin{table}[t]
\centering
\caption{Full-split generation results. ``Drop'' is the fraction with any later-budget
Token-F1 decrease above 2 score points. Paired-bootstrap 95\% CIs are shown for oracle gaps.}
\label{tab:generation}
\resizebox{\linewidth}{!}{%
\begin{tabular}{llrcccc}
\toprule
\textbf{Dataset} & \textbf{Model / grid} & $n$ & \textbf{Best fixed} & \textbf{Oracle} &
\textbf{Gap [95\% CI]} & \textbf{Drop} \\
\midrule
MLVU generation & Qwen2.5, text/16/64f & 418 & 33.0 & 35.7 & $+2.7\,[2.3,3.1]$ & 43.1\% \\
& Qwen3, text/16/64f & 418 & 33.3 & 37.0 & $+3.7\,[3.2,4.2]$ & 49.0\% \\
& LLaVA, text/8/32f & 418 & 34.9 & 38.0 & $+3.1\,[2.7,3.6]$ & 53.6\% \\
\midrule
AVSD turn 10 & Qwen2.5, text/16/64f & 1,787 & 30.2 & 35.0 & $+4.8\,[4.3,5.2]$ & 36.3\% \\
& Qwen3, text/16/64f & 1,787 & 33.5 & 37.3 & $+3.8\,[3.4,4.2]$ & 29.2\% \\
\bottomrule
\end{tabular}%
}
\end{table}

\begin{table}[t]
\centering
\caption{Token-F1 oracle gaps by generation regime (score points on a 0--100 scale).
``Long'' denotes the prespecified $\geq30$-minute MLVU QA subset; ``Long summary'' denotes
$\geq100$ reference words. Dashes indicate an unevaluated model--task combination.}
\label{tab:generation_strata_main}
\resizebox{0.96\linewidth}{!}{%
\begin{tabular}{lcccccc}
\toprule
\textbf{Model} & \textbf{MLVU QA} & \textbf{Summary} & \textbf{Long QA} &
\textbf{Long summary} & \textbf{AVSD yes/no} & \textbf{AVSD other} \\
\midrule
Qwen2.5 & +3.9 & +1.6 & +3.0 & +2.8 & +4.9 & +4.7 \\
Qwen3   & +5.5 & +1.8 & +3.0 & +2.6 & +3.7 & +4.0 \\
LLaVA   & +3.9 & +2.3 & +3.3 & +2.8 & -- & -- \\
\bottomrule
\end{tabular}%
}
\end{table}

\paragraph{Open-ended QA and summarization.}
MLVU contains 201 sub-scene questions and 217 video summaries. Positive Token-F1 oracle
gaps hold separately in both groups: QA/summary gaps are $+3.9/+1.6$ points for Qwen2.5,
$+5.5/+1.8$ for Qwen3, and $+3.9/+2.3$ for LLaVA. ROUGE-L is positive in every
model--task cell. The largest visual budget usually improves the mean, yet item-wise maxima
remain higher and later-budget regressions coexist with those gains. At a stricter five-point
drop threshold, regression remains $29.7$--$34.7\%$. Improved aggregate quality therefore
coexists with budget-specific per-item optima.

The empirical signal lies in the magnitude and uncertainty of the oracle gap.
Table~\ref{tab:generation_strata_main} shows positive gaps in every evaluated task group; all
six paired-bootstrap lower bounds for the long-video and long-summary strata are above zero.
Token-F1 and ROUGE-L also agree in direction for every model--task cell.

The split includes videos up to 117.1 minutes and summaries up to 242 reference words. On a
predefined $\geq30$-minute subset ($n{=}40$) and a $\geq100$-word summary subset
($n{=}130$), all three models retain positive score gaps with paired-bootstrap lower bounds
above zero (Appendix~\ref{app:longform}), extending the pattern to long videos and responses.

\paragraph{Controlled dialogue generation.}
For AVSD validation, turns 1--9 use the same reference history at every budget and only turn
10 is generated. This prevents history branching from confounding the current-turn visual
budget. Both models improve in average Token-F1 from text to $64f$, while the item oracle
remains $3.8$--$4.8$ points higher. The effect holds in both yes/no and non-yes/no groups and
under ROUGE-L (Appendix~\ref{app:avsd}). Holding history fixed and omitting audio isolates
the contribution of current-turn visual evidence.

These experiments broaden the response-trajectory evidence beyond option-letter flips while
preserving a task-appropriate contract: binary confusion for MCQA, continuous
complementarity and score regression for generation.

\section{What Changes the Trajectories?}
\label{sec:robustness}
\label{sec:negative}

We extend the controlled intervention beyond frame count and test four additional axes on all
786 Qwen2.5 V1-short items (Table~\ref{tab:robustness}). Every
grid retains a positive item-oracle gap. In particular, uniform and random sampling are almost
tied in aggregate ($73.92\%$ versus $74.05\%$) while exchanging 18 correct-to-wrong and
19 wrong-to-correct items. This is the same cancellation pattern under a fixed frame and pixel
budget.

\begin{table}[t]
\centering
\caption{Setup interventions on Qwen2.5-VL-7B, Video-MME v1 short. Pixel values are per
frame. Raw and cache are independently executed six-point grids.}
\label{tab:robustness}
\resizebox{\linewidth}{!}{%
\begin{tabular}{lccccl}
\toprule
\textbf{Intervention grid} & \textbf{Best fixed} & \textbf{Oracle} & \textbf{Gap} &
\textbf{Confusion} & \textbf{Additional diagnostic} \\
\midrule
Sampling: uniform/random/dense-start, 64f--151K & 74.05 & 77.74 & +3.69 & -- & 11.2\% policy-sensitive \\
Resolution: 64f, 76K/151K/235K/360K & 74.55 & 80.03 & +5.47 & -- & all four budgets uniquely useful \\
Iso-compute: 32f--302K/64f--151K/128f--76K & 75.32 & 81.55 & +6.23 & -- & 247/254/285 lowest-cost winners \\
Raw video: text/16/32/64/128/256f & 76.46 & 85.24 & +8.78 & 19.21\% & historical visuals exactly reproduced \\
1-fps cache: text/16/32/64/128/256f & 74.81 & 83.59 & +8.78 & 15.90\% & separate end-to-end execution \\
\bottomrule
\end{tabular}%
}
\end{table}

\begin{figure}[t]
\centering
\includegraphics[width=0.78\linewidth]{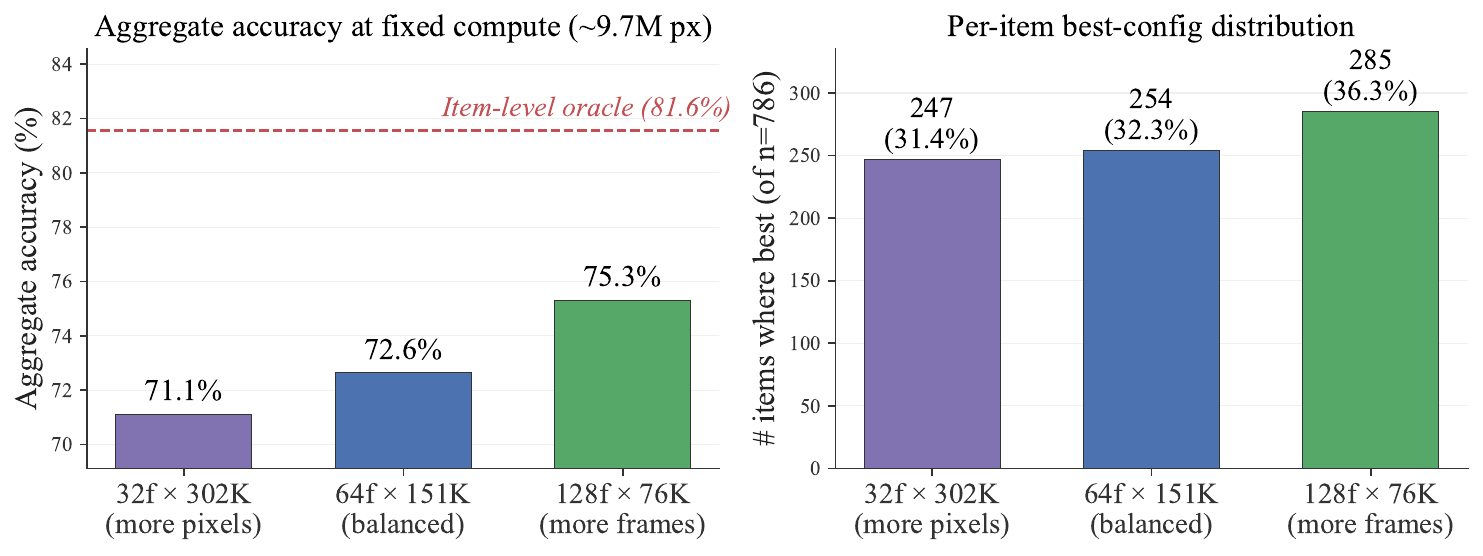}
\caption{\textbf{Iso-compute heterogeneity at approximately 9.7M total input pixels.}
$128f{\times}76$K wins on average, but $32f{\times}302$K, $64f{\times}151$K, and
$128f{\times}76$K are the per-item optimum for 247, 254, and 285 items. The item oracle is
$81.6\%$, $+6.2$ points above the best fixed allocation.}
\label{fig:isobudget}
\end{figure}

The matched-total-pixel intervention is especially informative because it holds the nominal
visual input area approximately fixed while reallocating it between temporal and spatial
coverage. Each allocation is optimal for hundreds of items (Figure~\ref{fig:isobudget}),
showing that both temporal and spatial allocation matter.

\subsection{Population-Level Replication, Item-Level Protocol Dependence}

A fresh raw-video rerun exactly reproduces predictions and correctness for all 786 items in
each historical visual cell. We then independently execute a true-cache six-point grid. Both
raw and cache grids yield $+8.78$ points of full-grid headroom; their difference is 0.00
points with 95\% CI $[-1.78,+1.91]$. The exact oracle accuracy and confusion rate differ,
however. Across protocols, full-grid confusion labels agree on $87.0\%$ of items,
text-to-128f overwrite on $96.3\%$, and trajectory class on $80.4\%$. Population-level
cancellation and complementarity replicate across protocols, while item membership tracks the
execution path. The release records both levels through explicit provenance tags.

\subsection{Intervention-Linked Failure Hypotheses}
\label{sec:mechanisms}

\paragraph{Sampling accounts for a measurable subset.}
Matched-grid confused items are sampling-sensitive at $30/113=26.5\%$, compared with
$58/673=8.6\%$ among other items (odds ratio 3.83, Fisher $p=4.75\times10^{-7}$).
Alternative sampling recovers 20/69 ($29.0\%$) terminal-$64f$ regressions; 49 persist under
all three policies. Evidence composition is therefore a controlled contributor to the observed
regressions.

\paragraph{Observed pathways.}
A deterministic 12-case frame audit identifies recurring, testable pathways. In temporal
counting cases, relevant events are separated or occluded and all policies retain the same
wrong count. In salient-cue cases, an OCR value from an earlier moment competes with the
queried later event. In sparse-referent cases, dominant scene content overwhelms a briefly
visible target. These observations connect behavior to visible evidence and supply testable
hypotheses for larger causal studies. Contact sheets and selection rules are released in
Appendix~\ref{app:mechanism_audit}.

\paragraph{Simple low-level statistics are weak predictors.}
Across 786 questions and 262 videos, none of eight motion/intensity proxies---duration,
adjacent-frame change summaries, first--last change, luminance, or spatial gradient---passes
BH-FDR ($|\rho|\leq0.078$, $|\delta_{\mathrm{Cliff}}|\leq0.087$). A multivariate
question-side audit finds no independent counting or numeric-option effect after permutation
correction. These diagnostics narrow the mechanism search beyond simple one-feature proxies.

\begin{table}[t]
\centering
\caption{Mechanism evidence from the controlled sampling intervention and diagnostic audits.}
\label{tab:mechanism_evidence}
\small
\setlength{\tabcolsep}{4pt}
\begin{tabular}{p{0.27\linewidth}p{0.28\linewidth}p{0.37\linewidth}}
\toprule
\textbf{Diagnostic} & \textbf{Estimate} & \textbf{Supported interpretation} \\
\midrule
Policy sensitivity & 88/786 items; policy oracle $+3.69$ pts & Frame selection changes
which items succeed even at fixed frame/pixel budgets. \\
Enrichment in confused set & 26.5\% vs. 8.6\%; OR 3.83 & Sampling-linked evidence
composition is disproportionately common among confused items. \\
Terminal-$64f$ recovery & 20/69 recovered; 49/69 persistent & Sampling changes a substantial
subset, while most regressions persist across all policies. \\
Motion/intensity proxies & 0/8 pass BH-FDR; $|\rho|\leq0.078$ & Duration and simple visual
dynamics are weak one-feature predictors. \\
Early-to-late predictability & slope $r=-0.16$; early selector 72.9\% & Cheap trajectory
signals reveal little of the later item-specific response. \\
\bottomrule
\end{tabular}
\end{table}

\subsection{Why the Oracle Is Hard to Reach}

Early and late frame-budget slopes correlate only $r=-0.16$ ($r^2\approx2.6\%$). A
selector restricted to text/$16f$/$32f$ features reaches $72.9\%$; a selector that observes
all 11 configurations reaches $77.9\%$ but requires executing the complete grid. Wrong
high-budget predictions on confused items can also remain confident. Together these results
locate the routing bottleneck: the decisive signal often appears only at the later
configuration.

\section{Artifact and Operational Implications}
\label{sec:implications}

\paragraph{A reusable audit contract.}
The release contains approximately 0.13M per-item model outputs, binary and continuous
trajectory labels, aggregate summaries, raw/cache and sampling provenance, and scripts for
recomputing every metric. A new model should begin with the common
$\{\text{text},16f,32f,64f\}$ grid, derive model-specific labels, and extend to higher
budgets only when its context permits. Released confusing-item IDs should be treated as
model-conditioned labels: most items in the cross-model confusion-set union are
model-exclusive, even though the overlap exceeds a matched-size random baseline.

\paragraph{A reference cascade.}
On Qwen2.5 V1 short, a standard margin cascade over $16f\!\rightarrow\!32f\!\rightarrow
128f$ matches fixed-$128f$ accuracy at 87.4 shared frames on average and lowers measured
wall-clock latency by $28.2\%$ when intermediate computation is reused. Without sharing it
is $13.7\%$ slower than fixed $128f$. Table~\ref{tab:cascade} separates deployable policies
from retrospective upper bounds.

\begin{table}[t]
\centering
\caption{Reference policies on Qwen2.5-VL-7B, V1 short.}
\label{tab:cascade}
\label{tab:main}
\begin{tabular}{lccc}
\toprule
\textbf{Policy} & \textbf{Accuracy} & \textbf{Average frame cost} & \textbf{Deployable} \\
\midrule
Fixed $128f$ & 76.5\% & 128.0 & \cmark \\
Confidence cascade & 76.5\% & 87.4 & \cmark, with sharing \\
Oracle-margin cascade & 78.8\% & 58.0 & \xmark \\
Full-trajectory selector & 77.9\% & 128.8 & \xmark \\
11-config item oracle & 87.9\% & -- & \xmark \\
\bottomrule
\end{tabular}
\end{table}

The deployment pattern transfers beyond the anchor model. Without per-model tuning, one
agreement rule---stop when $16f$ and $32f$ predict the same answer, otherwise evaluate
$64f$---retains accuracy close to the matched-grid best fixed policy for all three additional
V1-short models (Table~\ref{tab:cross_model_cascade}), demonstrating comparable
compute--accuracy audits across architectures.

\begin{table}[t]
\centering
\caption{Cross-model reference cascade with one untuned agreement rule. ``Gap'' is relative
to each model's matched-grid best fixed accuracy; frame cost assumes shared sampled frames.}
\label{tab:cross_model_cascade}
\begin{tabular}{lcccc}
\toprule
\textbf{Model} & \textbf{Accuracy} & \textbf{Gap} & \textbf{Avg. frames} & \textbf{Early stop} \\
\midrule
Qwen3-VL-8B    & 77.1\% & $-1.0$ pts & 37.7 & 82.2\% \\
InternVL3-8B   & 71.1\% & $-0.7$ pts & 36.1 & 87.3\% \\
InternVL3.5-8B & 73.8\% & $-0.4$ pts & 36.4 & 86.4\% \\
\bottomrule
\end{tabular}
\end{table}

\paragraph{A measurable routing objective.}
The response matrix supports a four-part evaluation of adaptive policies: fixed-budget
accuracy, average visual cost, measured latency, and the directional rescued/harmed flows.
The shared-compute cascade matches fixed-$128f$ accuracy while reducing average frames from
128.0 to 87.4; the untuned cross-model rule stops $82.2$--$87.3\%$ of items at
36.1--37.7 average frames while remaining within $0.4$--$1.0$ accuracy points of each
model's matched-grid best fixed policy. Reporting these quantities together distinguishes
genuine compute savings from policies that merely stop early on easy items.

These results suggest a minimum reporting standard for visual-budget scaling. First, publish
the aggregate curve together with paired rescued/harmed flows to interpret near-zero net
changes. Second, report both each model's full supported grid and the shared grid used for
cross-model comparison, isolating behavioral differences from architectural context ceilings.
Third, treat a change in decoder, cache, sampler, resolution, or input size as a new protocol
cell and preserve its provenance instead of pooling labels. For
generation, report per-item continuous scores under named metrics. These additions are inexpensive
once per-item outputs are retained, and they remain useful even when the downstream system
never deploys adaptive inference.

\section{Conclusion}
\label{sec:conclusion}
\label{sec:discussion}

Paired budget trajectories reveal the item-level dynamics hidden by aggregate Video LLM
scaling curves. Across five open models, four MCQA splits, open-ended QA and summarization,
and controlled dialogue generation, fixed visual budgets leave substantial configuration
complementarity. On the matched MCQA grid, oracle headroom spans $8.8$--$18.9$ points and
$12.5$--$25.5\%$ of items undergo harmful higher-budget transitions. The same structure
persists across frame count, resolution, sampling, temporal--spatial allocation, and raw/cache
execution.

The study turns scaling evaluation from a sequence of means into an auditable response
matrix: report rescued and harmed flows, compare models on matched grids, and preserve
protocol provenance. Our released 0.13M per-item records and analysis code support these
audits, while the cascade results demonstrate their value for compute-aware inference.
Interactive dialogue, human-preference metrics, and causal model-internal interventions are
natural next steps.

\clearpage
\subsection*{AI use statement}
Generative AI tools assisted with experiment planning, code implementation and debugging,
statistical analysis, result interpretation, literature organization, and manuscript drafting
and editing. The authors inspected the underlying data and execution artifacts, tested the
analysis code, enforced per-cell completeness checks, and manually verified all reported
claims and citations. The tools were not used as evaluators of the paper's primary results.
The authors take responsibility for the final content, including all AI-assisted text, code,
and analyses.

\subsection*{Ethics statement}
This work evaluates frozen models on existing public benchmarks and releases derived model
outputs and annotations rather than source media. We collected no new human-participant
data and redistribute no videos, frames, audio, subtitles, question text, or model
checkpoints. Users must obtain source assets from their original providers and follow the
corresponding terms. The released tags describe model behavior and do not encode personal
attributes of individuals appearing in videos. Appendix~\ref{app:ethics} documents source
governance, licensing boundaries, intended uses, and limitations.

\subsection*{Reproducibility statement}
The supplementary artifact contains per-item records, derived labels, protocol provenance,
analysis scripts, and cached/raw inference adapters. Appendix~\ref{app:artifact_full}
specifies the schema and release contract; Appendix~\ref{app:inference} records inference
settings; and Appendix~\ref{app:cache_validation} reports protocol-replication checks. Every
official cell is admitted only when item counts match the dataset exactly and missing,
duplicate, and error counts are zero.

\bibliography{references}

@misc{bai2025qwen25vltechnicalreport,
  title={Qwen2.5-{VL} Technical Report},
  author={Shuai Bai and Keqin Chen and Xuejing Liu and Jialin Wang and Wenbin Ge and Sibo Song and Kai Dang and Peng Wang and Shijie Wang and Jun Tang and Humen Zhong and Yuanzhi Zhu and Mingkun Yang and Zhaohai Li and Jianqiang Wan and Pengfei Wang and Wei Ding and Zheren Fu and Yiheng Xu and Jiabo Ye and Xi Zhang and Tianbao Xie and Zesen Cheng and Hang Zhang and Zhibo Yang and Haiyang Xu and Junyang Lin},
  year={2025},
  eprint={2502.13923},
  archivePrefix={arXiv},
  primaryClass={cs.CV},
  url={https://arxiv.org/abs/2502.13923},
}

@misc{bai2026qwen3vl,
  title={Qwen3-{VL} Technical Report},
  author={Bai, Shuai and Cai, Yuxuan and Chen, Ruizhe and others},
  year={2025},
  eprint={2511.21631},
  archivePrefix={arXiv},
  primaryClass={cs.CV},
  url={https://arxiv.org/abs/2511.21631}
}

@article{zhu2025internvl3,
  title={{InternVL3}: Exploring Advanced Training and Test-Time Recipes for Open-Source Multimodal Models},
  author={Zhu, Jinguo and Wang, Weiyun and Chen, Zhe and Liu, Zhaoyang and Ye, Shenglong and Gu, Lixin and Tian, Hao and Duan, Yuchen and Su, Weijie and Shao, Jie and others},
  journal={arXiv preprint arXiv:2504.10479},
  year={2025}
}

@misc{zhu2026internvl35,
  title={{InternVL3.5}: Advancing Open-Source Multimodal Models in Versatility, Reasoning, and Efficiency},
  author={Wang, Weiyun and Gao, Zhangwei and Gu, Lixin and Pu, Hengjun and Cui, Long and others},
  year={2025},
  eprint={2508.18265},
  archivePrefix={arXiv},
  primaryClass={cs.CV},
  url={https://arxiv.org/abs/2508.18265}
}

@article{llava_next_video_2024,
  title={{LLaVA}-{NeXT}: A Strong Zero-Shot Video Understanding Model},
  author={Zhang, Yuanhan and Li, Bo and Liu, haotian and Lee, Yong jae and Gui, Liangke and Fu, Di and Feng, Jiashi and Liu, Ziwei and Li, Chunyuan},
  journal={LLaVA Blog},
  year={2024},
  url={https://llava-vl.github.io/blog/2024-04-30-llava-next-video/}
}

@article{li2023videochat,
  title={{VideoChat}: Chat-Centric Video Understanding},
  author={Li, KunChang and He, Yinan and Wang, Yi and Li, Yizhuo and Wang, Wenhai and Luo, Ping and Wang, Yali and Wang, Limin and Qiao, Yu},
  journal={arXiv preprint arXiv:2305.06355},
  year={2023}
}

@article{zhang2023videollama,
  title={Video-{LLaMA}: An Instruction-tuned Audio-Visual Language Model for Video Understanding},
  author={Zhang, Hang and Li, Xin and Bing, Lidong},
  journal={arXiv preprint arXiv:2306.02858},
  year={2023}
}

@article{zhang2024llavavideo,
  title={Video Instruction Tuning with Synthetic Data},
  author={Zhang, Yuanhan and Wu, Jinming and Li, Wei and Li, Bo and Ma, Zejun and Liu, Ziwei and Li, Chunyuan},
  journal={arXiv preprint arXiv:2410.02713},
  year={2024}
}

@article{li2024llavaonevision,
  title={{LLaVA}-{OneVision}: Easy Visual Task Transfer},
  author={Li, Bo and Zhang, Yuanhan and Guo, Dong and Zhang, Renrui and Li, Feng and Zhang, Hao and Zhang, Kaichen and Zhang, Peiyuan and Li, Yanwei and Liu, Ziwei and Li, Chunyuan},
  journal={arXiv preprint arXiv:2408.03326},
  year={2024}
}

@article{ye2024mplugowl3,
  title={{mPLUG-Owl3}: Towards Long Image-Sequence Understanding in Multi-Modal Large Language Models},
  author={Ye, Jiabo and Xu, Haiyang and Liu, Haowei and Hu, Anwen and Yan, Ming and Qian, Qi and Zhang, Ji and Huang, Fei and Zhou, Jingren},
  journal={arXiv preprint arXiv:2408.04840},
  year={2024}
}

@article{shu2025videoxl,
  title={Video-{XL}: Extra-Long Vision Language Model for Hour-Scale Video Understanding},
  author={Shu, Yan and Liu, Zheng and Zhang, Peitian and Qin, Minghao and Zhou, Junjie and Liang, Zhengyang and Huang, Tiejun and Zhao, Bo},
  journal={arXiv preprint arXiv:2409.14485},
  year={2024}
}

@inproceedings{fu2025video,
  title={Video-{MME}: The First-Ever Comprehensive Evaluation Benchmark of Multi-Modal {LLM}s in Video Analysis},
  author={Fu, Chaoyou and Dai, Yuhan and Luo, Yongdong and Li, Lei and Ren, Shuhuai and Zhang, Renrui and Wang, Zihan and Zhou, Chenyu and Shen, Yunhang and Zhang, Mengdan and others},
  booktitle={Proceedings of the {IEEE/CVF} Conference on Computer Vision and Pattern Recognition ({CVPR})},
  year={2025}
}

@article{videomme_v2_2026,
  title={Video-{MME}-v2: Towards the Next Stage in Benchmarks for Comprehensive Video Understanding},
  author={Fu, Chaoyou and others},
  journal={arXiv preprint arXiv:2604.05015},
  year={2026}
}

@inproceedings{mlvu_2024,
  title={{MLVU}: Benchmarking Multi-task Long Video Understanding},
  author={Zhou, Junjie and Shu, Yan and Zhao, Bo and Wu, Boya and Liang, Zhengyang and Xiao, Shitao and Qin, Minghao and Yang, Xi and Xiong, Yongping and Zhang, Bo and Huang, Tiejun and Liu, Zheng},
  booktitle={Proceedings of the {IEEE/CVF} Conference on Computer Vision and Pattern Recognition ({CVPR})},
  pages={13691--13701},
  year={2025}
}

@inproceedings{li2024mvbench,
  title={{MVBench}: A Comprehensive Multi-Modal Video Understanding Benchmark},
  author={Li, Kunchang and Wang, Yali and He, Yinan and Li, Yizhuo and Wang, Yi and Liu, Yi and Wang, Zun and Xu, Jilan and Chen, Guo and Luo, Ping and Wang, Limin and Qiao, Yu},
  booktitle={Proceedings of the {IEEE/CVF} Conference on Computer Vision and Pattern Recognition ({CVPR})},
  year={2024}
}

@inproceedings{liu2024tempcompass,
  title={{TempCompass}: Do Video {LLM}s Really Understand Videos?},
  author={Liu, Yuanxin and Li, Shicheng and Liu, Yi and Wang, Yuxiang and Ren, Shuhuai and Li, Lei and Chen, Sishuo and Sun, Xu and Hou, Lu},
  booktitle={Findings of the Association for Computational Linguistics ({ACL})},
  year={2024}
}

@inproceedings{mangalam2023egoschema,
  title={{EgoSchema}: A Diagnostic Benchmark for Very Long-form Video Language Understanding},
  author={Mangalam, Karttikeya and Akshulakov, Raiymbek and Malik, Jitendra},
  booktitle={Advances in Neural Information Processing Systems ({NeurIPS})},
  year={2023}
}

@inproceedings{liu2024mmbench,
  title={{MMBench}: Is Your Multi-Modal Model an All-Around Player?},
  author={Liu, Yuan and Duan, Haodong and Zhang, Yuanhan and Li, Bo and Zhang, Songyang and Zhao, Wangbo and Yuan, Yike and Wang, Jiaqi and He, Conghui and Liu, Ziwei and Chen, Kai and Lin, Dahua},
  booktitle={Proceedings of the European Conference on Computer Vision ({ECCV})},
  year={2024}
}

@article{zhang2025vps,
  title={Video Parallel Scaling: Aggregating Diverse Frame Subsets for Video {LLM}s},
  author={Chung, Hyungjin and Nam, Hyelin and Kim, Jiyeon and Go, Hyojun and Park, Byeongjun and Kim, Junho and Lee, Joonseok and Ha, Seongsu and Kim, Byung-Hoon},
  journal={arXiv preprint arXiv:2509.08016},
  year={2025}
}

@article{li2025afp,
  title={Less is More: Token-Efficient Video-QA via Adaptive Frame-Pruning and Semantic Graph Integration},
  author={Wang, Shaoguang and Guo, Weiyu and Chen, Ziyang and Xu, Yijie and Hu, Xuming and Xiong, Hui},
  journal={arXiv preprint arXiv:2508.03337},
  year={2025}
}

@article{huang2025framesampling,
  title={Frame Sampling Strategies Matter: A Benchmark for Small Vision-Language Models},
  author={Brkic, Marija and Razzouki, Anas Filali and Tevissen, Yannis and Guetari, Khalil and El Yacoubi, Mounim A.},
  journal={arXiv preprint arXiv:2509.14769},
  year={2025}
}

@inproceedings{zohar2025apollo,
  title={{Apollo}: An Exploration of Video Understanding in Large Multimodal Models},
  author={Zohar, Orr and Wang, Xiaohan and Dubois, Yann and Mehta, Nikhil and Xiao, Tong and Hansen-Estruch, Philippe and Yu, Licheng and Wang, Xiaofang and Juefei-Xu, Felix and Zhang, Ning and Yeung-Levy, Serena and Xia, Xide},
  booktitle={Proceedings of the {IEEE/CVF} Conference on Computer Vision and Pattern Recognition ({CVPR})},
  year={2025}
}

@inproceedings{li2025f16,
  title={Improving {LLM} Video Understanding with $16$ Frames Per Second},
  author={Li, Yixuan and Tang, Changli and Zhuang, Jimin and Yang, Yudong and Sun, Guangzhi and Zhang, Chao and Li, Wei and Ma, Zejun},
  booktitle={Proceedings of the International Conference on Machine Learning ({ICML})},
  year={2025}
}

@inproceedings{zhang2025flashvid,
  title={{FlashVID}: Efficient Video Large Language Models via Training-free Tree-based Spatiotemporal Token Merging},
  author={Fan, Ziyang and Chen, Keyu and Xing, Ruilong and Li, Yulin and Jiang, Li and Tian, Zhuotao},
  booktitle={International Conference on Learning Representations ({ICLR})},
  year={2026},
  eprint={2602.08024},
  archivePrefix={arXiv},
  url={https://arxiv.org/abs/2602.08024}
}

@article{you2025focus,
  title={{FOCUS}: Efficient Keyframe Selection for Long Video Understanding},
  author={Zhu, Zirui and Xu, Hailun and Luo, Yang and Liu, Yong and Sarkar, Kanchan and Yang, Zhenheng and You, Yang},
  journal={arXiv preprint arXiv:2510.27280},
  year={2025}
}

@inproceedings{luo2025framevoyager,
  title={Frame-Voyager: Learning to Query Frames for Video Large Language Models},
  author={Yu, Sicheng and Jin, Chengkai and Wang, Huanyu and Chen, Zhenghao and Jin, Sheng and Zuo, Zhongrong and Xu, Xiaolei and Sun, Zhenbang and Zhang, Bingni and Wu, Jiawei and Zhang, Hao and Sun, Qianru},
  booktitle={International Conference on Learning Representations ({ICLR})},
  year={2025}
}

@inproceedings{fan2025air,
  title={{AIM}: Adaptive Inference of Multi-Modal {LLM}s via Token Merging and Pruning},
  author={Zhong, Yiwu and Liu, Zhuoming and Li, Yin and Wang, Liwei},
  booktitle={Proceedings of the {IEEE/CVF} International Conference on Computer Vision ({ICCV})},
  year={2025},
  eprint={2412.03248},
  archivePrefix={arXiv},
  url={https://arxiv.org/abs/2412.03248}
}

@inproceedings{hu2025mllmframe,
  title={{M-LLM} Based Video Frame Selection for Efficient Video Understanding},
  author={Hu, Kai and Gao, Feng and Nie, Xiaohan and Zhou, Peng and Tran, Son and Neiman, Tal and Wang, Lingyun and Shah, Mubarak and Hamid, Raffay and Yin, Bing and Chilimbi, Trishul},
  booktitle={Proceedings of the {IEEE/CVF} Conference on Computer Vision and Pattern Recognition ({CVPR})},
  year={2025}
}

@article{chen2025breakingdown,
  title={Breaking Down Video {LLM} Benchmarks: Knowledge, Spatial Perception, or True Temporal Understanding?},
  author={Feng, Bo and Lai, Zhengfeng and Li, Shiyu and Wang, Zizhen and Wang, Simon and Huang, Ping and Cao, Meng},
  journal={arXiv preprint arXiv:2505.14321},
  year={2025}
}

@inproceedings{swayamdipta2020cartography,
  title={Dataset Cartography: Mapping and Diagnosing Datasets with Training Dynamics},
  author={Swayamdipta, Swabha and Schwartz, Roy and Lourie, Nicholas and Wang, Yizhong and Hajishirzi, Hannaneh and Smith, Noah A. and Choi, Yejin},
  booktitle={Proceedings of the Conference on Empirical Methods in Natural Language Processing ({EMNLP})},
  year={2020}
}

@inproceedings{baldock2021depth,
  title={Deep Learning Through the Lens of Example Difficulty},
  author={Baldock, Robert J.~N. and Maennel, Hartmut and Neyshabur, Behnam},
  booktitle={Advances in Neural Information Processing Systems ({NeurIPS})},
  year={2021}
}

@inproceedings{yue2024cascade,
  title={Large Language Model Cascades with Mixture of Thought Representations for Cost-Efficient Reasoning},
  author={Yue, Murong and Zhao, Jie and Zhang, Min and Du, Liang and Yao, Ziyu},
  booktitle={International Conference on Learning Representations ({ICLR})},
  year={2024}
}

@article{chen2024cascadeaware,
  title={Cascade-Aware Training of Language Models},
  author={Wang, Congchao and Augenstein, Sean and Rush, Keith and Jitkrittum, Wittawat and Narasimhan, Harikrishna and Rawat, Ankit Singh and Menon, Aditya Krishna and Go, Alec},
  journal={arXiv preprint arXiv:2406.00060},
  year={2024}
}

@inproceedings{jitkrittum2025speculative,
  title={Faster Cascades via Speculative Decoding},
  author={Narasimhan, Harikrishna and Jitkrittum, Wittawat and Rawat, Ankit Singh and Kim, Seungyeon and Gupta, Neha and Menon, Aditya Krishna and Kumar, Sanjiv},
  booktitle={International Conference on Learning Representations ({ICLR})},
  year={2025},
  eprint={2405.19261},
  archivePrefix={arXiv},
  url={https://arxiv.org/abs/2405.19261}
}

@inproceedings{chen2025fastvlm,
  title={{FREE}: Fast and Robust Vision Language Models with Early Exits},
  author={Bajpai, Divya Jyoti and Hanawal, Manjesh Kumar},
  booktitle={Findings of the Association for Computational Linguistics ({ACL})},
  year={2025}
}

@inproceedings{deer_vla_2024,
  title={{DeeR-VLA}: Dynamic Inference of Multimodal Large Language Models for Efficient Robot Execution},
  author={Yue, Yang and Wang, Yulin and Kang, Bingyi and Han, Yizeng and Wang, Shenzhi and Song, Shiji and Feng, Jiashi and Huang, Gao},
  booktitle={Advances in Neural Information Processing Systems ({NeurIPS})},
  year={2024}
}

@inproceedings{dtoma_2025,
  title={{DToMA}: Training-free Dynamic Token Manipulation for Long Video Understanding},
  author={Yuan, Bowen and You, Sisi and Bao, Bing-Kun},
  booktitle={Proceedings of the International Joint Conference on Artificial Intelligence ({IJCAI})},
  year={2025}
}

@inproceedings{wei2022cot,
  title={Chain-of-Thought Prompting Elicits Reasoning in Large Language Models},
  author={Wei, Jason and Wang, Xuezhi and Schuurmans, Dale and Bosma, Maarten and Ichter, Brian and Xia, Fei and Chi, Ed and Le, Quoc V. and Zhou, Denny},
  booktitle={Advances in Neural Information Processing Systems ({NeurIPS})},
  year={2022}
}

@inproceedings{wang2023self,
  title={Self-Consistency Improves Chain of Thought Reasoning in Language Models},
  author={Wang, Xuezhi and Wei, Jason and Schuurmans, Dale and Le, Quoc V. and Chi, Ed H. and Narang, Sharan and Chowdhery, Aakanksha and Zhou, Denny},
  booktitle={International Conference on Learning Representations ({ICLR})},
  year={2023}
}

@inproceedings{snell2024scaling,
  title={Scaling {LLM} Test-Time Compute Optimally Can Be More Effective Than Scaling Model Parameters},
  author={Snell, Charlie and Lee, Jaehoon and Xu, Kelvin and Kumar, Aviral},
  booktitle={International Conference on Learning Representations ({ICLR})},
  year={2025}
}

@inproceedings{chen2024fastv,
  title={An Image is Worth $1/2$ Tokens After Layer $2$: Plug-and-Play Inference Acceleration for Large Vision-Language Models},
  author={Chen, Liang and Zhao, Haozhe and Liu, Tianyu and Bai, Shuai and Lin, Junyang and Zhou, Chang and Chang, Baobao},
  booktitle={Proceedings of the European Conference on Computer Vision ({ECCV})},
  year={2024}
}

@inproceedings{alamri2019audio,
  title     = {Audio Visual Scene-Aware Dialog},
  author    = {Alamri, Huda and Cartillier, Vincent and Das, Abhishek and Wang, Jue and
               Cherian, Anoop and Essa, Irfan and Batra, Dhruv and Marks, Tim K. and
               Hori, Chiori and Anderson, Peter and Lee, Stefan and Parikh, Devi},
  booktitle = {Proceedings of the IEEE/CVF Conference on Computer Vision and Pattern Recognition},
  pages     = {7558--7567},
  year      = {2019}
}

@article{lin2026stepaudio,
  title={Stepaudio 2.5 technical report},
  author={Lin, Bin and Zhao, Bo and Wu, Boyong and Yan, Chao and Wu, Chen and Yi, Cheng and Yao, Chengyuan and Liu, Daijiao and Tian, Fei and Tian, Feng and others},
  journal={arXiv preprint arXiv:2605.23463},
  year={2026}
}

@inproceedings{wang2023crosssinger,
  title={Crosssinger: A cross-lingual multi-singer high-fidelity singing voice synthesizer trained on monolingual singers},
  author={Wang, Xintong and Zeng, Chang and Chen, Jun and Wang, Chunhui},
  booktitle={2023 IEEE Automatic Speech Recognition and Understanding Workshop (ASRU)},
  pages={1--6},
  year={2023},
  organization={IEEE}
}

@article{sun2026muse,
  title={MUSE: A Multi-agent Framework for Unconstrained Story Envisioning via Closed-Loop Cognitive Orchestration},
  author={Sun, Wenzhang and Wang, Zhenyu and Hu, Zhangchi and Wang, Chunfeng and Li, Hao and Chen, Wei},
  journal={arXiv preprint arXiv:2602.03028},
  year={2026}
}

@inproceedings{chenbeyond,
  title={Beyond Logits: Coherent Hallucination Mitigation via Attention Contrastive Decoding},
  author={Chen, Yujia and Sun, Rui and Mai, Huayu and Li, Wangkai and He, Zhangyu and Wang, Bingzhou and Li, Aibing and SUN, Wenzhang and Zhang, Tianzhu},
  booktitle={Forty-third International Conference on Machine Learning}
}
\bibliographystyle{iclr2027_conference}

\appendix
\section{Extended Task-Format Experiments}
\label{app:task_formats}

\subsection{MLVU Open-Ended Generation}
\label{app:mlvu_generation}

We evaluate all 418 MLVU generation items: 201 sub-scene questions and 217 video
summaries. Qwen2.5 and Qwen3 use text/$16f$/$64f$; LLaVA uses text/$8f$/$32f$.
Generation is greedy with at most 256 new tokens. Table~\ref{tab:mlvu_task_breakdown}
reports deterministic Token-F1; ROUGE-L gives a positive oracle gap in every row. A frozen
Qwen2.5-VL-32B rubric judge is retained as a secondary analysis after exact-answer and direct-
contradiction controls, but it is not the official MLVU GPT-4 evaluator and is not used for the
primary claim.

\begin{table}[h]
\centering
\caption{MLVU Token-F1 oracle gaps by task group. Gaps are score points on a 0--100 scale.}
\label{tab:mlvu_task_breakdown}
\resizebox{0.82\linewidth}{!}{%
\begin{tabular}{llrc}
\toprule
\textbf{Model} & \textbf{Task} & $n$ & \textbf{Oracle gap} \\
\midrule
Qwen2.5 & sub-scene QA & 201 & +3.9 \\
& summary & 217 & +1.6 \\
Qwen3 & sub-scene QA & 201 & +5.5 \\
& summary & 217 & +1.8 \\
LLaVA & sub-scene QA & 201 & +3.9 \\
& summary & 217 & +2.3 \\
\bottomrule
\end{tabular}%
}
\end{table}

The gaps are recomputed from the same 418 full outputs used in
Table~\ref{tab:generation}; machine-readable means, item scores, and bootstrap samples are
included in the artifact.

\subsection{Long-Video and Longer-Summary Strata}
\label{app:longform}

The $\geq30$-minute sub-scene subset contains 40 questions (median 63.1 minutes; maximum
117.1), and the $\geq100$-reference-word summary subset contains 130 items (median 126;
maximum 242 words). Table~\ref{tab:longform} reports Token-F1 oracle gaps. These are
coverage analyses: task type is coupled with each stratum, so they do not identify duration or
answer length as an independent cause.

\begin{table}[h]
\centering
\caption{Long-video and longer-summary Token-F1 oracle gaps with 95\% paired-bootstrap CIs.}
\label{tab:longform}
\begin{tabular}{lcc}
\toprule
\textbf{Model} & $\geq30$ min ($n{=}40$) & $\geq100$ words ($n{=}130$) \\
\midrule
Qwen2.5 & $+3.0\,[1.3,4.8]$ & $+2.8\,[1.6,4.0]$ \\
Qwen3   & $+3.0\,[1.2,5.0]$ & $+2.6\,[1.4,4.0]$ \\
LLaVA   & $+3.3\,[1.3,4.7]$ & $+2.8\,[1.5,4.1]$ \\
\bottomrule
\end{tabular}
\end{table}

At stricter $\geq60$-minute ($n{=}22$) and $\geq150$-word ($n{=}34$) thresholds, all six
point estimates remain positive ($+1.2$ to $+4.1$ points); five bootstrap lower bounds are
positive and one reaches zero.

\subsection{AVSD Fixed-History Current-Turn Generation}
\label{app:avsd}

We use all 1,787 ten-turn dialogues in AVSD@DSTC7 validation. Turns 1--9 are fixed to the
same ground-truth history for every budget, and only turn 10 is generated. Audio and the
optional questioner summary are omitted, isolating current-turn visual evidence. All six
inference cells and six secondary-judge cells contain 1,787 clean rows. Table~\ref{tab:avsd}
reports Token-F1; ROUGE-L oracle gaps are $+4.6/+3.7$ points for Qwen2.5/Qwen3.

\begin{table}[h]
\centering
\caption{AVSD Token-F1 by answer group. ``Means'' lists text/$16f$/$64f$.}
\label{tab:avsd}
\resizebox{0.88\linewidth}{!}{%
\begin{tabular}{llrcccc}
\toprule
\textbf{Model} & \textbf{Group} & $n$ & \textbf{Means} & \textbf{Best fixed} & \textbf{Oracle} & \textbf{Gap} \\
\midrule
Qwen2.5 & all & 1,787 & 29.2/30.2/30.2 & 30.2 & 35.0 & +4.8 \\
& yes/no & 1,127 & 30.8/31.6/31.6 & 31.6 & 36.5 & +4.9 \\
& other & 660 & 26.4/27.6/27.8 & 27.8 & 32.5 & +4.7 \\
Qwen3 & all & 1,787 & 31.6/32.7/33.5 & 33.5 & 37.3 & +3.8 \\
& yes/no & 1,127 & 33.2/34.6/35.1 & 35.1 & 38.8 & +3.7 \\
& other & 660 & 28.9/29.5/30.7 & 30.7 & 34.7 & +4.0 \\
\bottomrule
\end{tabular}%
}
\end{table}

For a later-drop threshold of 1/2/5 score points, regression rates are
40.1/36.3/30.4\% for Qwen2.5 and 32.3/29.2/23.5\% for Qwen3. The direction is stable to
the threshold, while the rate decreases as expected.

\section{Protocol Replication and Mechanism Audits}

\subsection{Full Raw-Video and Cache Grids}
\label{app:raw_cache_full}

The historical Qwen2.5 V1-short visual cells used raw-video decoding. A fresh raw-video rerun
reproduces all five visual JSONLs exactly in predictions, correctness, and row content. We
then run an independent cache grid. Table~\ref{tab:raw_cache_full} reports the two full grids.

\begin{table}[h]
\centering
\caption{Full six-point raw/cache replication. Accuracy values are percentages.}
\label{tab:raw_cache_full}
\resizebox{0.9\linewidth}{!}{%
\begin{tabular}{lcccccc|cccc}
\toprule
& \multicolumn{6}{c|}{\textbf{Accuracy by configuration}} & \textbf{Best} & \textbf{Oracle} & \textbf{Gap} & \textbf{Confusion} \\
\textbf{Protocol} & T & 16 & 32 & 64 & 128 & 256 & \textbf{fixed} & & & \\
\midrule
Raw & 37.53 & 65.90 & 68.96 & 72.65 & 76.46 & 75.83 & 76.46 & 85.24 & +8.78 & 19.21 \\
Cache & 37.40 & 64.50 & 70.74 & 73.92 & 74.81 & 74.81 & 74.81 & 83.59 & +8.78 & 15.90 \\
\bottomrule
\end{tabular}%
}
\end{table}

Full-grid headroom differs by 0.00 points (95\% CI $[-1.78,+1.91]$). Exact trajectory
vectors agree on 68.83\% of items, confusion on 87.02\%, text-to-128f overwrite on 96.31\%,
and trajectory class on 80.41\%. We therefore report protocol-specific labels and use the
comparison as end-to-end replication, not as a cache-only causal estimate.

\subsection{Sampling and Structured Mechanism Audit}
\label{app:mechanism_audit}

At fixed $64f$/$151$K, uniform, random-seed-42, and dense-start sampling obtain
73.92\%, 74.05\%, and 70.48\%. Across policies, 88/786 items are sampling-sensitive and
the policy oracle is 77.74\%, $+3.69$ points over the best fixed policy. Confused items are
3.83 times more likely to be sampling-sensitive; alternative policies recover 20/69 terminal
regressions.

The structured audit selects 12 non-duplicate cases by prespecified behavioral signatures:
four sampling-recovered, four sampling-persistent, two counting regressions, and two OCR
regressions. Representative cases include: (i) separated flag or toast occurrences that all
policies undercount; (ii) a visible 5:59 clock competing with the queried later 6:00 event;
and (iii) a rare camera-holder overwhelmed by dominant rugby/referee frames. The artifact
contains the case table, frame-contact sheets, and selection manifest.

Across all 786 questions and 262 videos, none of eight low-level motion/intensity features
passes BH-FDR. A six-feature question-side logistic audit also finds no positive risk feature
that survives permutation BH-FDR after accounting for correlated counting and numeric-option
indicators. These negative results rule out a simple one-feature explanation without excluding
semantic integration or attention-based mechanisms.

\section{Additional Experimental Details}
\label{app:details}
\label{app:artifact_details}

\subsection{Released Artifact: Detailed Contents and Schema}
\label{app:artifact_full}

The artifact has four components, all under a single release directory.

\paragraph{(A) Per-item correctness labels.}
For each (model, benchmark split, configuration) triple we evaluated, we release one row per item in CSV form with the schema:
\texttt{sample\_id, task\_type, video\_length\_sec, model, split, config, frame\_count, pixel\_count, predicted\_option, ground\_truth\_option, correct, source\_tag, protocol\_note} and optional timing fields.
The provenance fields identify raw/cache execution, sampling policy, input-size fallback, and
the official-clean source manifest. This is the lowest-level record from which every other
label is deterministically derived. The release contains approximately 0.13M rows after
applying the per-model coverage matrix (Appendix~\ref{app:coverage}).

\paragraph{(B) Derived per-item annotations.}
A second CSV joins on \texttt{(model, split, sample\_id)} and adds: \texttt{is\_visually\_confused} (binary), \texttt{is\_text\_overwritten} (binary), \texttt{trajectory\_class} (one of $10$ patterns; Appendix~\ref{app:taxonomy}), \texttt{best\_config} (cheapest configuration that answers the item correctly under the model's full grid; \texttt{None} if no configuration does), and \texttt{matched\_grid\_best\_config} (same, restricted to the matched $4$-config sub-grid \{text-only, $16f$, $32f$, $64f$\}).

\paragraph{(C) Aggregated derived data.}
For each (model, split) cell, JSON files report: (i)~the scaling curve; (ii)~the $2\!\times\!2$ text-vs-video overwrite matrix and its task-wise breakdown; (iii)~the trajectory taxonomy distribution and its task-type cross-tabulation; (iv)~the bidirectional churn matrix at every adjacent budget transition; (v)~the matched-grid headroom and confusion rate. These are the inputs to every figure and table in the paper; figure-generation scripts are also released.

\paragraph{(D) Evaluation pipeline (code).}
The package contains raw-video and cached-frame inference adapters, a one-time 1-fps JPEG
cache builder, completeness/merge checks, label derivation, figure-data generation, and the
reference cascade stack. Sampling policy and protocol provenance are explicit command-line
arguments and output fields.

\paragraph{Reproducibility.}
Per-item outputs use greedy decoding under recorded precision and attention settings. Official
cells are selected by a manifest and admitted only when dataset and merged counts match and
missing, duplicate, and error counts are zero. The raw-video anchor cells reproduce exactly
under the same runner; independently executed cache cells are retained as separate protocol
records (Appendix~\ref{app:cache_validation}). A new Video LLM can be added by implementing
one inference adapter; earlier rows remain read-only.

\paragraph{License and distribution.}
Code is released under the MIT License and newly created labels and annotations under
CC-BY-4.0. These licenses do not override source-benchmark terms. We do \emph{not}
redistribute videos, extracted frames, audio, subtitles, question text, or checkpoints; users
must obtain source assets from the original providers. A Croissant 1.1 record documents
provenance, schema, licenses, intended uses, limitations, and sensitive-information boundaries.

\paragraph{Intended use and out-of-scope use.}
The artifact supports: reproduction of every empirical claim; development of new adaptive-inference methods (cascades, routers, early-exit predictors) along the frame-budget axis with a calibrated upper bound (the per-item oracle); benchmark-level analyses of item-level scaling heterogeneity in Video LLMs more broadly. The labels are \emph{not} intended for: claiming aggregate ``more frames hurt'' regressions on benchmarks beyond the four (model, split) cells we evaluated; training a model-agnostic ``difficult-item'' predictor without re-running cross-model overlap checks (the cross-model audit finds both shared and model-specific structure); ranking models against each other on raw confusion rates without applying the matched-grid normalization.

\paragraph{Limitations of the released labels.}
(i)~The configuration grid is non-uniform across models due to architectural ceilings.
(ii)~Exact item labels are protocol-conditioned; raw and cache outputs must not be pooled.
(iii)~Alternative sampling policies are available only for the Qwen2.5 V1-short intervention.
(iv)~Confidence/margin fields are available only for cells used by the cascade analysis.
(v)~Continuous generation scores are metric-dependent and are not binary correctness labels.

\subsection{Inference Configuration}
\label{app:inference}

MCQA experiments use BF16 precision, greedy decoding (temperature $= 0$), a maximum of
128 output tokens, and Flash Attention 2 where supported. Generation experiments allow up to
256 new tokens. Visual inputs follow the recorded per-cell protocol: either uniform decoding
from raw video or uniform subsampling from the 1-fps JPEG cache; the sampling intervention
additionally uses random-seed-42 and dense-start policies.
For Qwen2.5-VL-7B and Qwen3-VL-8B, spatial resolution is controlled via the \texttt{max\_pixels} parameter.
For InternVL3-8B and InternVL3.5-8B, \texttt{input\_size} is set to $448$ and the number of tiles per frame follows the official dynamic-image-size behavior for video inputs.
For LLaVA-NeXT-Video-7B-hf, we use the HF-converted checkpoint (\texttt{llava-hf/LLaVA-NeXT-Video-7B-hf}) and the \texttt{LlavaNextVideoProcessor} with its default resolution; the context limit ($4096$ tokens) restricts our evaluation to at most $32$ frames per item.

\subsection{Coverage Matrix}
\label{app:coverage}

Table~\ref{tab:coverage} summarizes the matched grid used for Table~\ref{tab:mcqa_grid}.
Every listed cell is official-clean: merged and dataset counts match exactly, with no missing,
duplicate, or error rows. Higher-budget and fallback cells are reported separately rather than
mixed into matched-grid comparisons.
The two InternVL-family V2-medium $64f$ cells use their clean zero-error reruns; earlier
incomplete high-memory attempts are excluded from both the table and released official labels.

\begin{table}[h]
\centering
\caption{Official matched-grid coverage for the four main models. All 16 model--split rows
contain text/$16f$/$32f$/$64f$.}
\label{tab:coverage}
\resizebox{0.72\textwidth}{!}{%
\begin{tabular}{llcccc}
\toprule
\textbf{Model} & \textbf{Split} & T & 16 & 32 & 64 \\
\midrule
Qwen2.5-VL-7B  & V1 short  & \cmark & \cmark & \cmark & \cmark \\
                & V1 medium & \cmark & \cmark & \cmark & \cmark \\
                & V2 medium & \cmark & \cmark & \cmark & \cmark \\
                & MLVU      & \cmark & \cmark & \cmark & \cmark \\
Qwen3-VL-8B    & V1 short  & \cmark & \cmark & \cmark & \cmark \\
                & V1 medium & \cmark & \cmark & \cmark & \cmark \\
                & V2 medium & \cmark & \cmark & \cmark & \cmark \\
                & MLVU      & \cmark & \cmark & \cmark & \cmark \\
InternVL3-8B   & V1 short  & \cmark & \cmark & \cmark & \cmark \\
                & V1 medium & \cmark & \cmark & \cmark & \cmark \\
                & V2 medium & \cmark & \cmark & \cmark & \cmark \\
                & MLVU      & \cmark & \cmark & \cmark & \cmark \\
InternVL3.5-8B & V1 short  & \cmark & \cmark & \cmark & \cmark \\
                & V1 medium & \cmark & \cmark & \cmark & \cmark \\
                & V2 medium & \cmark & \cmark & \cmark & \cmark \\
                & MLVU      & \cmark & \cmark & \cmark & \cmark \\
\bottomrule
\end{tabular}%
}
\end{table}

Qwen2.5 and Qwen3 additionally include clean $128f$ cells and selected $256f$ cells.
InternVL $128f$ results use a separately tagged $\texttt{input\_size}{=}224$ fallback and are
not substituted for the standard-$448$ matched cells. LLaVA uses its context-safe
text/$8f$/$16f$/$32f$ grid and is reported as a third-family replication rather than inserted
into Table~\ref{tab:mcqa_grid}.

\subsection{Cached Pipeline and Sanity Validation}
\label{app:cache_validation}

Cached cells extract frames at 1 FPS with \texttt{ffmpeg}, store JPEGs at quality $q{=}2$,
and uniformly subsample the requested budget. Raw cells decode directly from source video.
The historical Qwen2.5 anchor cells are raw; later coverage cells carry explicit source tags.
We never merge item labels across those paths.

The initial single-cell check compared cached and raw execution at $64f$:

\begin{itemize}[nosep,leftmargin=*]
  \item \textbf{Qwen2.5-VL-7B}: online $72.65\%$ vs.\ cached $73.92\%$, absolute delta $1.27$ points, per-item agreement $91.86\%$ ($n=786$).
  \item \textbf{InternVL3-8B}: online $70.99\%$ vs.\ cached $69.59\%$, absolute delta $1.40$ points, per-item agreement $93.77\%$ ($n=786$).
\end{itemize}

This check motivated the full six-point replication in Appendix~\ref{app:raw_cache_full}.
Both full grids retain $+8.78$ points of headroom, while oracle accuracy, confusion rate, and
item membership differ. Consequently, the paper treats population-level replication and
single-item label invariance as separate questions.

\paragraph{Item-level robustness audit (stratified subset).}
Before completing the full grids, we also ran a stratified $n{=}150$ audit ($50$
stable-correct, $30$ monotonic-increasing, $30$ visually confused, $20$ text-overwritten,
and $20$ random fill). Table~\ref{tab:cache_audit_items} records this historical diagnostic.
Because it is stratified and precedes the full replication, it is not used to estimate
population-wide invariance; the full-grid numbers are authoritative.

\begin{table}[h]
\centering
\caption{Item-level cache-vs-online robustness audit on a stratified $n{=}150$ subset of Video-MME short. Each row reports the agreement rate of a single per-item label between the two pipelines.}
\label{tab:cache_audit_items}
\resizebox{0.6\textwidth}{!}{%
\begin{tabular}{lcc}
\toprule
\textbf{Per-item label} & \textbf{Agreement (\%)} & \textbf{Cohen's $\kappa$} \\
\midrule
Final correctness @ $64f$ & $91.9$ & $0.84$ \\
Best-config label (over 11 configs) & $87.3$ & $0.79$ \\
Visual-confusion tag (binary) & $94.0$ & $0.87$ \\
Text-overwrite tag (binary) & $96.7$ & $0.91$ \\
Trajectory class (10-way) & $90.7$ & $0.86$ \\
Oracle gap on subset (pts) & $\Delta{=}0.4$ & --- \\
\bottomrule
\end{tabular}%
}
\end{table}

\subsection{Visual Confusion Definition}
\label{app:confusion_def}

We define the compute cost ordering over configurations as follows.
For configurations varying only in frame count (at fixed resolution), cost is proportional to frame count.
For configurations varying only in resolution (at fixed frame count), cost is proportional to resolution.
For mixed comparisons, we use total pixel count (frames $\times$ resolution) as the cost measure.
An item is visually confused if it is correct at \emph{any} lower-cost configuration and incorrect at \emph{any} higher-cost configuration.

\subsection{Trajectory Taxonomy Construction}
\label{app:taxonomy}

Each item on the short split yields a binary correctness vector of length $6$ (text-only, $16f$, $32f$, $64f$, $128f$, $256f$, all at $151$K).
We assign items to pattern classes by inspecting monotonicity and the location of transitions:
\begin{itemize}[nosep,leftmargin=*]
  \item \textbf{Always correct} / \textbf{Always wrong}: all six entries agree.
  \item \textbf{Monotonic increasing}: at least one $0{\to}1$ transition, no $1{\to}0$ transition.
  \item \textbf{Monotonic decreasing}: at least one $1{\to}0$ transition, no $0{\to}1$.
  \item \textbf{Inverted-U}: correct on an interior stretch, wrong at both ends.
  \item \textbf{U-shape}: wrong on an interior stretch, correct at both ends.
  \item \textbf{Text-dominant}: text-only correct but at least one higher-budget wrong.
  \item \textbf{Other non-monotonic}: oscillating, late-recovery, etc.
\end{itemize}
The medium-split taxonomy uses a $4$-point vector (text-only, $16f$, $64f$, $128f$) and only retains patterns that are identifiable at length $4$; subtle patterns like oscillating collapse into \textit{other non-monotonic}.

\paragraph{Full taxonomy distribution.}
\begin{table}[h]
\centering
\small
\caption{Full $10$-class trajectory taxonomy on V1 short and V1 medium for Qwen2.5-VL-7B. Non-monotonic + text-dominant items together are the structural origin of visual confusion.}
\resizebox{0.9\textwidth}{!}{%
\begin{tabular}{lcc}
\toprule
\textbf{Pattern} & \textbf{V1 short ($n{=}786$)} & \textbf{V1 medium ($n{=}639$)} \\
\midrule
Always correct & $26.6\%$ & $27.2\%$ \\
Always wrong & $14.8\%$ & $21.9\%$ \\
Monotonic increasing & $39.4\%$ & $32.9\%$ \\
Monotonic decreasing & $0.6\%$ & $1.6\%$ \\
Inverted-U & $4.2\%$ & $6.4\%$ \\
U-shape & $3.8\%$ & $5.6\%$ \\
Text dominant (text-only correct, later wrong) & $2.5\%$ & $2.8\%$ \\
Other non-monotonic (oscillating, late-recovery) & $8.1\%$ & $1.6\%$ \\
\midrule
\textbf{Non-monotonic total (structural confusion)} & $\mathbf{19.2\%}$ & $\mathbf{18.0\%}$ \\
\bottomrule
\end{tabular}%
}
\end{table}

\paragraph{Visual-confusion rate by task type (V1 short, anchor model).}
\begin{table}[h]
\centering
\small
\caption{Per-task visual confusion rate on Qwen2.5-VL-7B $\times$ V1 short across the $11$-config grid (top $8$ of $26$ task types).}
\resizebox{0.58\textwidth}{!}{%
\begin{tabular}{lcc}
\toprule
\textbf{Task type} & \textbf{Confusion rate (\%)} & \textbf{Confused items} \\
\midrule
Counting Problem & $36.1$ & $39$ \\
OCR Problems & $24.5$ & $12$ \\
Action Recognition & $22.0$ & $24$ \\
Object Reasoning & $19.7$ & $13$ \\
Object Recognition & $19.5$ & $29$ \\
Spatial Reasoning & $15.4$ & $4$ \\
Spatial Perception & $14.8$ & $4$ \\
Attribute Perception & $14.0$ & $15$ \\
\bottomrule
\end{tabular}%
}
\end{table}

\subsection{Feature Importance Analysis}
\label{app:features}
\label{app:feature_importance}

\begin{figure}[h]
\centering
\includegraphics[width=0.6\textwidth]{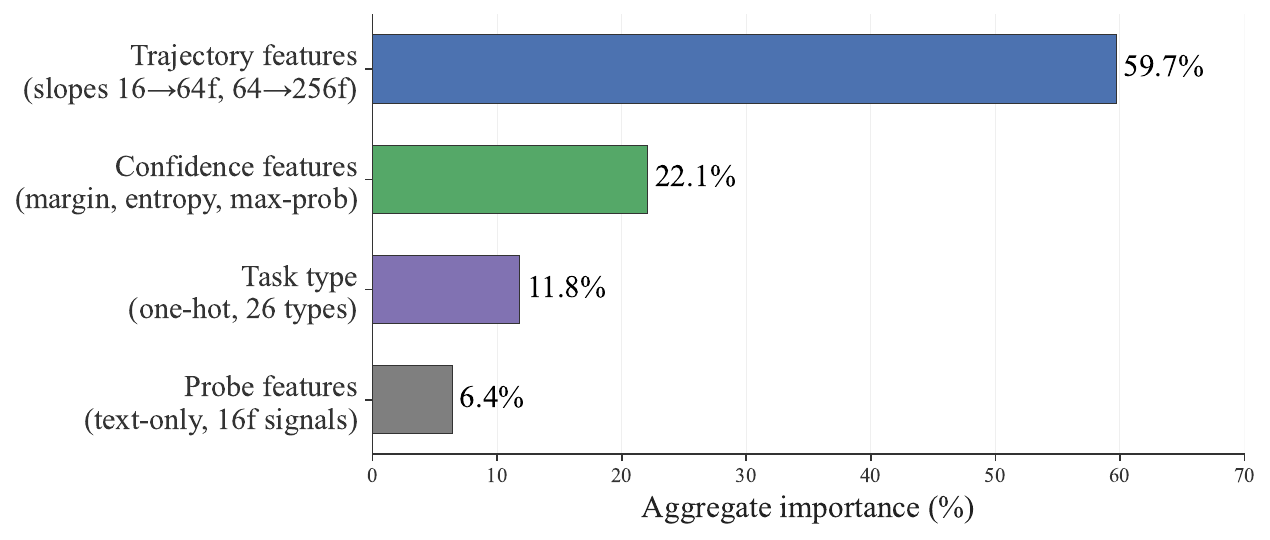}
\caption{Feature importance for the oracle GBM selector (Qwen2.5-VL-7B on short). Trajectory features (requiring the full scaling curve) dominate, accounting for $59.7\%$ of total importance.}
\label{fig:feature_importance_full}
\end{figure}

The oracle selector uses $14$ numeric features plus one-hot encoded task type.
Feature groups and their aggregate importance:
\begin{itemize}[nosep,leftmargin=*]
  \item \textbf{Trajectory features} ($59.7\%$): slope $64{\to}256f$ ($35.6\%$), slope $16{\to}64f$ ($24.1\%$)
  \item \textbf{Confidence features} ($22.1\%$): answer margin, entropy, max option probability
  \item \textbf{Task type} ($11.8\%$): one-hot encoding of $26$ Video-MME task types
  \item \textbf{Probe features} ($6.4\%$): text-only correctness, $16f$ accuracy
\end{itemize}

Removing trajectory features drops accuracy from $77.9\%$ to $72.3\%$ ($-5.6$ points), while removing all other feature groups individually causes drops of $<1$ point.

\subsection{LongVideoBench Details}
\label{app:lvb}

LVB evaluation uses $50$ items from the test split, each evaluated at two stages (stage~1: $0.5$ FPS, $151$K; stage~2: $2.0$ FPS, $235$K). Stage~$1$ accuracy $64.0\%$, Stage~$2$ $62.0\%$, two-stage oracle $70.0\%$, confused items $4$ ($8.0\%$), helped items $3$ ($6.0\%$), stage agreement $78.0\%$.
Stage~$2$ margins are almost entirely saturated (mean margin $=1.0$), precluding margin-based cascade analysis on LVB.
The benchmark is retained as qualitative reference; the Video-MME medium split (\S\ref{sec:findings}) is our primary cross-benchmark evidence.

\section{Cross-Model Details}
\label{app:crossmodel}

\subsection{Cross-model cascade reference baselines}
\label{app:crossmodel_cascade}

A single agreement-based cascade template, without per-model tuning, transfers qualitatively
to additional models (Table~\ref{tab:cross_model_cascade}). These reference cascades remain
within 0.4--1.0 points of the corresponding matched-grid best-fixed accuracy while using
substantially fewer shared frames. We interpret them as evidence that the artifact supports
practical compute--accuracy analysis across architectures, not as optimized adaptive-inference
policies. Cheap early stages resolve many easy items, while later stages handle the
visually-dependent residue.

\subsection{Qwen3-VL-8B and InternVL3.5-8B Short Curves}
\label{app:newgen_curves}

Table~\ref{tab:newgen_short} reports the full per-configuration accuracy for the two next-generation models on Video-MME short.

\begin{table}[h]
\centering
\caption{Per-configuration accuracy on Video-MME short for Qwen3-VL-8B (6 configs, up to $256f$) and InternVL3.5-8B (4 configs, up to $64f$). Both models exhibit substantial between-configuration variance consistent with the item-level heterogeneity story.}
\label{tab:newgen_short}
\resizebox{0.7\textwidth}{!}{%
\begin{tabular}{lcc}
\toprule
\textbf{Config} & \textbf{Qwen3-VL-8B (\%)} & \textbf{InternVL3.5-8B (\%)} \\
\midrule
text-only & 43.9 & 42.5 \\
$16f$     & 69.6 & 70.6 \\
$32f$     & 74.8 & 73.9 \\
$64f$     & 78.1 & 74.2 \\
$128f$    & \textbf{79.6} & --- \\
$256f$    & \textbf{79.6} & --- \\
\midrule
\rowcolor{oraclecolor!10}Item-level oracle & $88.7$ & $84.7$ \\
Oracle headroom (pts) & $+9.0$ & $+10.6$ \\
Confusion rate (\%) & $16.7$ & $14.1$ \\
\bottomrule
\end{tabular}%
}
\end{table}

Several qualitative observations from this table:
\begin{itemize}[nosep,leftmargin=*]
  \item Qwen3-VL-8B ties itself at $128f$ and $256f$ ($79.64\%$ exact match), indicating the short-split cached $1$-FPS regime is saturated by $128f$ for this model. Unlike Qwen2.5-VL-7B (which \emph{regresses} at $256f$), Qwen3-VL-8B saturates---likely reflecting the larger $262$K context absorbing the extra visual tokens without attention-distribution strain.
  \item InternVL3.5-8B's accuracy rises monotonically through $64f$ at a decelerating rate; the model has not reached saturation within its evaluated range.
  \item After the V1-short consistency audit (Appendix~\ref{app:qwen3_audit}), Qwen3-VL-8B's confusion rate is $16.7\%$ over its six-configuration full grid and $16.3\%$ on the four-configuration matched grid. The matched-grid value aligns with InternVL3.5-8B ($14.1\%$) and Qwen2.5-VL-7B ($15.9\%$); its larger matched-grid headroom on V1 medium and MLVU remains visible in Table~\ref{tab:crossmodel_summary}.
\end{itemize}

On medium, Qwen3-VL-8B yields text-only $42.6\%$, $16f$ $47.3\%$, $64f$ $47.3\%$, $128f$ $54.6\%$, while InternVL3.5-8B yields text-only $40.1\%$ and $64f$ $63.2\%$.
The Qwen3-VL-8B medium curve notably shows $16f \approx 64f$ followed by a jump at $128f$, a pattern not observed on short.

\subsection{Qwen3-VL-8B Video-MME V1 Short Audit}
\label{app:qwen3_audit}

We performed an additional consistency audit for Qwen3-VL-8B on Video-MME V1 short after detecting an anomalous preliminary scaling curve.
The anomalous run showed a large regression at high frame budgets that was inconsistent with the other models and splits.
We therefore re-ran every Qwen3-VL-8B V1-short visual configuration using the same cleaned cached-frame pipeline and prompt templates as the rest of the matrix.

The audited Qwen3-VL-8B V1-short curve is shown below.
\begin{center}
\small
\begin{tabular}{lcccccc}
\toprule
Configuration & text-only & $16f$ & $32f$ & $64f$ & $128f$ & $256f$ \\
\midrule
Accuracy (\%) & $43.89$ & $69.59$ & $74.81$ & $78.12$ & $79.64$ & $79.64$ \\
\bottomrule
\end{tabular}
\end{center}
The corrected curve is monotonic up to saturation, and the $128f$ and $256f$ predictions are identical at the item level ($626/786$ correct).
All V1-short Qwen3-VL-8B statistics reported in the paper---oracle headroom, visual confusion, text overwrite, trajectory taxonomy, and matched-grid metrics---are computed from this audited run.

We also audited the existing Qwen3-VL-8B runs on V1 medium, V2, and MLVU using stratified subsets and the same cleaned runner path.
Minimum per-configuration correctness agreement was $98.0\%$ on V1 medium, $100\%$ on V2, and $100\%$ on MLVU.
We therefore retain the original full-grid outputs for those splits.

\paragraph{InternVL3-8B V1-short consistency audit.}
We also repeated the InternVL3-8B V1-short $64f$ run under the standard $448$ input-size protocol.
The repeated run yields $71.37\%$ compared with the earlier $69.59\%$ measurement, a $+1.78$ point difference.
At the item level, correctness agreement is $95.42\%$, with $36$ items flipping correctness between the two runs.
The aggregate difference is fully explained by these sparse item-level flips.
We therefore treat this as benign run-to-run variation rather than a protocol drift, and use the audited run consistently in the reported matrix.

\subsection{Cross-Model Confused-Item Overlap}
\label{app:jaccard}

Over the shared short configurations (text-only, $16f$, $32f$, $64f$), Qwen2.5-VL-7B has $125$ visually confused items and InternVL3-8B has $98$. Their intersection contains $33$ items and their union $190$, giving a Jaccard overlap of $\mathbf{17.4\%}$.
Our null preserves the two set sizes and randomizes item membership over the shared $786$-item universe. The exact matched-size random-set expectation is $7.5\%$ Jaccard with a $95\%$ interval of $[4.2\%,10.9\%]$. The observed overlap is $\mathbf{2.30\times}$ this expectation ($p=2.2\times10^{-6}$, exact upper-tail probability).

\textbf{Interpretation (consistent with main-text \S\ref{sec:findings}).} The above-chance overlap indicates a shared benchmark or task component: some items expose scaling sensitivity in both models. At the same time, $157/190=82.6\%$ of union items are exclusive to one model's confused set. The result combines a shared task component with substantial model-specific membership.

\subsection{InternVL3.5-8B: Effective Single-Pass Frame Ceiling}
\label{app:i35_highbudget}

The InternVL3.5-8B model is architected with a $32{,}768$-token context window set during its supervised fine-tuning stage~\citep{zhu2026internvl35}.
Under standard video inference with $\texttt{max\_num}{=}1$ (one tile per frame), each frame contributes exactly $256$ visual tokens after the architectural pixel-unshuffle---a product of $(448/14)^2 / 4 = 256$ for $448{\times}448$ input with ViT patch $14$ and $0.5$ downsample ratio.
The official HuggingFace inference example uses this configuration.

The effective single-pass frame ceiling is therefore:
\[
N_{\max} \;=\; \left\lfloor \frac{32{,}768 - T_{\text{prompt}}}{256 + T_{\text{frame-prefix}}} \right\rfloor \;\approx\; 120 \text{ frames},
\]
where $T_{\text{prompt}}$ ($\sim\!400$--$600$ tokens) covers system prompt, question, options, and formatting, and $T_{\text{frame-prefix}}\!\approx\! 6$ accounts for the per-frame \texttt{Frame\{i\}:\,<image>} marker.
We independently verified this budget: $64f$ consumes $\sim\!52\%$ of the context, $96f$ consumes $\sim\!77\%$, and $128f$ mechanically exceeds the $32{,}768$ budget by design.

Our early $128f$ runs on InternVL3.5-8B produced a $256{:}64$ tokens-per-frame mismatch between the prompt-side placeholder count (standard image-mode) and the encoder output, interacting with a conservative tokenizer default \texttt{model\_max\_length}$=14{,}588$ (well below the architectural $32$K).
Bringing these into alignment---standard $256$-token layout plus explicit tokenizer override to $32{,}768$---makes budgets below the effective ceiling arithmetically feasible. We use the standard-input $64f$ cell in the matched grid; no incomplete higher-frame attempt contributes to any reported metric.
Accordingly, the InternVL3.5-8B main-grid coverage in this paper is capped at $64f@448$ (consistent with the official technical report~\citep{zhu2026internvl35}), and all oracle, confusion, and taxonomy analyses for this model are computed over \{text-only, $16f$, $32f$, $64f\}$.

\paragraph{Why standard $128f@448$ is invalid under this protocol.}
We attempted the standard $128f@448$ configuration, but it produced no valid MLVU item rows. The binding limit is architectural rather than simply aggregate GPU memory: $128 \times 256 = 32{,}768$ visual tokens already equals InternVL3.5-8B's full SFT-stage context window before prompt and per-frame marker overhead. Tensor or pipeline parallelism can shard weights but does not extend the trained context budget. Sequence-parallel methods could execute longer sequences only under a different inference protocol and would still extrapolate beyond the trained positions. We therefore reject single-pass $128f@448$ rather than treat it as a missing standard-grid result.

\paragraph{The context-safe input-size fallback we report instead.}
We report $128f$ at $\texttt{input\_size}{=}224$ as a context-safe fallback: at this resolution each frame contributes $(224/14)^2 / 4 = 64$ tokens, so $128f \times 64 = 8{,}192$ visual tokens, well within the $32$K budget. The MLVU $128f@224$ run yields $65.71\%$ ($1{,}426/2{,}170$, $0$ errors). This number is reported as a separately tagged data point and is not used as a substitute for the standard-input-size protocol; it is included to give a concrete data point at $128$ frames under the strongest protocol that fits the architectural budget. Notably, $128f@224$ is \emph{below} the InternVL3.5 $64f@448$ accuracy ($69.08\%$), so for this model on MLVU, the practical scaling ceiling under our protocol is $64f$ at standard input size, not $128f$.

Notably, the official InternVL3.5 technical report~\citep{zhu2026internvl35} reports Video-MME, MMBench-Video, MLVU, and LongVideoBench results using up to $64$ frames per item, consistent with the standard-input matched-grid ceiling used here.

\section{Supplementary Findings: Resolution Surface, Iso-Pixel, and Sampling Robustness}
\label{app:supplementary_findings}

This appendix extends the main empirical narrative of \S\ref{sec:findings} along the spatial-resolution and sampling-policy axes, with additional analyses of the phenomenon, mechanisms, and operational use case.

\subsection{Two-Dimensional Scaling Surface (Frame-by-Resolution)}
\label{app:2d_surface}

\begin{figure}[h]
\centering
\includegraphics[width=0.55\linewidth]{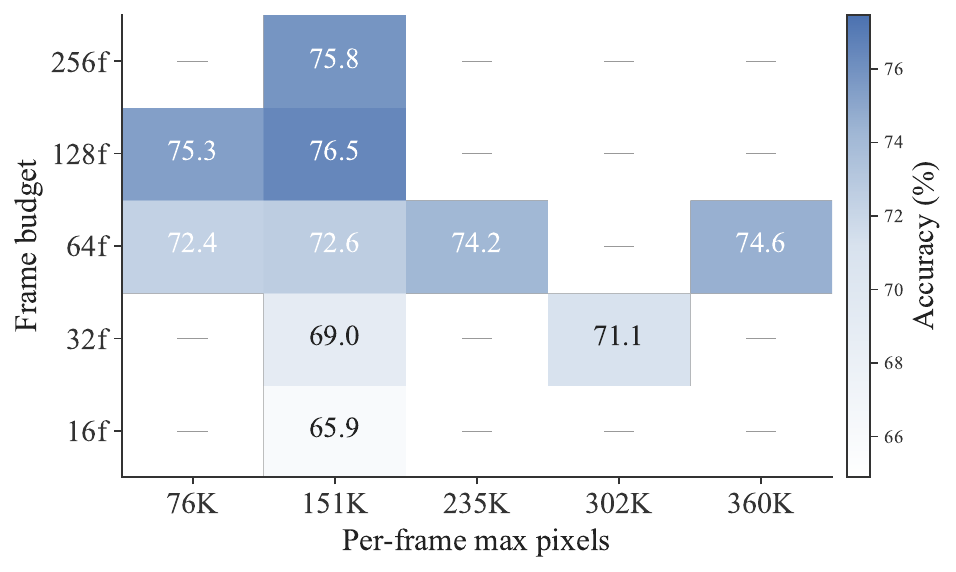}
\caption{Aggregate accuracy on the partially-filled frame$\times$resolution grid (Qwen2.5-VL-7B on short). Global optima cluster at high frame counts ($128f$), but per-item optima span the full surface. The iso-budget slice along the diagonal is promoted to the main body (Figure~\ref{fig:isobudget}).}
\label{fig:heatmap}
\end{figure}

The scaling heterogeneity reported in \S\ref{sec:findings} extends beyond the temporal axis into the spatial-resolution dimension.

\paragraph{Resolution heterogeneity at fixed frame count.}
At a fixed $64$ frames, we evaluate four resolutions on Qwen2.5-VL-7B $\times$ V1 short.
While aggregate accuracy increases monotonically ($72.4\% \to 74.6\%$), the per-item resolution oracle reaches $\mathbf{80.0\%}$, a $+5.5$-point gap over the best fixed resolution.
The best-resolution distribution is nearly uniform ($194$, $205$, $226$, $161$ items favor $76$K, $151$K, $235$K, $360$K respectively); the lowest resolution is optimal for nearly as many items as the highest---a direct counterexample to the assumption that higher resolution is universally beneficial.
We observe \emph{resolution confusion} paralleling frame confusion: $29$ items are correct at $76$K but wrong at $360$K, and $23$ at $151$K but wrong at $360$K.

\paragraph{Iso-budget analysis (frame $\leftrightarrow$ resolution trade-off).}
At a matched $\sim$9.7M total pixel budget, the per-item oracle across the three iso-budget configurations reaches $\mathbf{81.6\%}$, a $+6.2$-point headroom over the best fixed allocation ($128f{\times}76K$, $75.3\%$).
The best-allocation distribution ($247$, $254$, $285$) is remarkably balanced (Figure~\ref{fig:isobudget}), confirming that the temporal-vs-spatial trade-off is genuinely item-dependent.

\paragraph{Independent iso-pixel sweep at fixed $64f$.}
To isolate the resolution dimension under a clean full-benchmark protocol, we evaluate Qwen2.5-VL-7B on V1 short at fixed $64$ frames across three pixel budgets, each as a full $786$-item run: $50$K pixels yields $70.10\%$, $151$K pixels yields $73.92\%$, $200$K pixels yields $74.43\%$.
The $50\text{K} \to 151\text{K}$ jump is $+3.82$ points, while $151\text{K} \to 200\text{K}$ adds only $+0.51$ points---resolution is a genuine confound but its returns saturate beyond the $\sim$$151$K default scale.
The $151$K accuracy in this independently executed sweep ($73.92\%$) differs from the original $11$-configuration grid's $64f{\times}151K$ entry ($72.7\%$; \S\ref{sec:protocol}) because the sweep used the released final runner whereas the anchor grid predates that runner revision. We retain both protocol-tagged records rather than combine them; all within-grid comparisons use a single execution protocol consistently.

\subsection{Sampling-Strategy Robustness Ablation}
\label{app:sampling_robustness}

We test sampling-policy sensitivity on Qwen2.5-VL-7B $\times$ V1 short at fixed
$64f \times 151K$ pixels by replacing uniform sampling with two alternatives, each as a full
$786$-item run: \emph{dense-at-start} (concentrate all $64$ frames in the first $1/3$ of the
video) and \emph{random} (seed $42$).
Aggregate accuracies are uniform $73.92\%$, dense-at-start $70.48\%$, random $74.05\%$: random is statistically tied with uniform ($+0.13$ pt), while dense-at-start drops $3.44$ points by sacrificing temporal coverage.
Per-item correctness sets agree strongly between random and uniform (Jaccard $0.94$, $4.7\%$ of items flip), and only modestly less between dense-at-start and uniform (Jaccard $0.88$, $9.0\%$ flip).
Random sampling preserves the aggregate and per-item structure, while the systematic gap from dense-at-start quantifies the effect of reduced temporal coverage.

\subsection{Iso-budget Analysis}
\label{app:isobudget_analysis}

\noindent\textbf{Heterogeneity along frame$\times$resolution.}
The same phenomenon persists when we vary spatial resolution at matched compute (Figure~\ref{fig:isobudget}).
At a fixed $\sim$$9.7$M total-pixel budget, $32f{\times}302K$, $64f{\times}151K$, and $128f{\times}76K$ achieve aggregate accuracies of $71.1\%$, $72.6\%$, and $75.3\%$, respectively.
Yet each allocation is the per-item optimum for a large subset: $247$, $254$, and $285$ items.
The iso-budget item-level oracle reaches $\mathbf{81.6\%}$, a $+6.2$-pt gap over the best fixed allocation.
Thus, heterogeneity is not only about frame count; the temporal--spatial allocation itself is item-dependent.

\section{LLaVA-NeXT-Video-7B as Older-Generation Reference}
\label{app:llava_nextvideo}

We include LLaVA-NeXT-Video-7B-hf~\citep{llava_next_video_2024} as an older-generation architectural reference (trained in $2024$-$04$ on Vicuna-7B-v1.5).
Its $4096$-token context window restricts evaluation to $\{\text{text-only}, 8f, 16f, 32f\}$ on the short split at the default HF resolution:
text-only $28.4\%$, $8f$ $42.4\%$, $16f$ $43.6\%$, $32f$ $45.9\%$; best-fixed $45.9\%$, item-level oracle $57.8\%$, oracle headroom $+11.8$ points, visual confusion rate $16.2\%$.
LLaVA-NeXT-Video is substantially weaker than the main four models in absolute accuracy (consistent with its older training recipe and base LLM), but the oracle headroom and confusion rate fall squarely within the ranges reported for the main models, reinforcing that item-level scaling heterogeneity is not tied to a particular family or generation.

\paragraph{Full LLaVA replication on MLVU.}
A full $2{,}170$-item MLVU run under the same $\{\text{text-only}, 8f, 16f, 32f\}$ grid yields text-only $36.59\%$, $8f$ $45.30\%$, $16f$ $47.14\%$, $32f$ $48.02\%$; best-fixed $48.02\%$ ($32f$), item-level oracle $60.46\%$, oracle headroom $\mathbf{+12.4}$ \textbf{points}, visual confusion rate $17.7\%$ (Table~\ref{tab:crossmodel_summary}, bottom block).
LLaVA's MLVU absolute accuracy is the lowest in the cross-family set (Qwen2.5/Qwen3/InternVL3/InternVL3.5 reach $56$--$69\%$ best-fixed), but its oracle headroom and confusion rate lie within the same range as the main models, reinforcing that the per-item heterogeneity story is robust to architectural generation under a benchmark with materially different task composition from Video-MME.
LLaVA is not promoted to a main-matrix model because its $4$K context budget caps evaluation at $32f$ on MLVU, well below the $64f$/$128f$ ceiling at which the Qwen and InternVL families peak.

\section{Extended Cascade Details}
\label{app:cascade_details}

\subsection{Full Cascade Sweep (Qwen2.5-VL-7B on short)}

\begin{table}[h]
\centering
\caption{Full cascade sweep results. The best deployable point across all designs remains the frame-only $16f{\to}32f{\to}128f$ cascade.}
\label{tab:cascade_full}
\resizebox{0.7\textwidth}{!}{%
\begin{tabular}{lcccl}
\toprule
\textbf{Cascade Design} & \textbf{Accuracy (\%)} & \textbf{Cost} & \textbf{Stop Rate (\%)} & \textbf{Signal} \\
\midrule
\multicolumn{5}{l}{\textit{Deployable (margin-gated)}} \\
\quad $16f\to128f$ & 76.5 & 92.1 & 32.1 & margin \\
\quad $\mathbf{16f\to32f\to128f}$ & \textbf{76.5} & \textbf{87.4} & \textbf{37.5} & margin \\
\quad $32f\to128f$ & 76.5 & 96.8 & 32.5 & margin \\
\quad $64f\to128f$ & 76.5 & 105.2 & 35.6 & margin \\
\midrule
\multicolumn{5}{l}{\textit{Oracle-gated (upper bounds)}} \\
\quad $16f\to128f$ & 78.8 & 58.0 & 62.5 & $p_\text{correct}$ \\
\quad $32f_\text{302K}\to128f_\text{76K}$ & 78.9 & 84.4 & 68.2 & $p_\text{correct}$ \\
\bottomrule
\end{tabular}%
}
\end{table}

\subsection{Confusion Recovery by Cascade Variant}

Analyzing the oracle-gated cascade (using the true correctness signal as the gate), we find that it correctly protects $21$ of the $151$ visually confused items by stopping early, while incorrectly stopping only $3$ items, yielding a net benefit of $+18$ items.
The deployable margin-based cascade is more conservative: its primary benefit is compute reduction rather than confusion recovery, consistent with its moderate stop rate ($37.5\%$).

\subsection{Resolution-Aware and Mixed-Dimension Cascades}

We additionally evaluate cascades along the resolution dimension (\eg, $64f{\times}76K \to 64f{\times}235K$) and mixed-dimension designs (\eg, $32f{\times}302K \to 128f{\times}76K$).
While oracle-gated variants improve the upper bound ($78.9\%$ at cost $84.4$), the best \emph{deployable} cascade across all designs remains the frame-only $16f \to 32f \to 128f$ cascade, suggesting that frame-count changes produce larger and more reliable probability shifts than resolution changes.

\subsection{Cross-Model Cascade Details}
\label{app:crossmodel_cascade_details}

The cross-model cascade reference baselines are summarized in Table~\ref{tab:cross_model_cascade}.
The full sweep includes one additional Qwen3 variant: \emph{agree($16f$,$32f$)$\to 128f$}, which reaches $77.0\%$ at $49.1$ average frames with an $82.2\%$ stop rate.
This trades slightly more compute for essentially the same accuracy as the $\to 64f$ variant.
We use the same untuned agreement rule for every model. The result supports the operational
relevance of the artifact across model families.

\section{Runtime and Memory Measurements}
\label{app:runtime}

We report wall-clock latency, decode time, model forward time, and GPU memory usage on a $100$-item stratified subset of Video-MME short (stratified by task-type, seed $42$).
All measurements use a single NVIDIA L20X (144 GB) GPU, Qwen2.5-VL-7B with BF16 precision, Flash Attention 2, the offline cached frame pipeline, and greedy decoding of at most $128$ output tokens.
Wall-clock is end-to-end per-item including cache loading, processor invocation, model forward, and answer extraction.

\begin{table}[h]
\centering
\caption{Wall-clock runtime and peak GPU memory on Qwen2.5-VL-7B $\times$ Video-MME short $100$-item stratified subset. ``vs.\ $128f$'' is the relative wall-clock change with respect to the fixed $128f$ baseline (negative numbers indicate a speedup).}
\label{tab:runtime}
\resizebox{\textwidth}{!}{%
\begin{tabular}{lcccccc}
\toprule
\textbf{Config} & \textbf{Acc.\ (\%)} & \textbf{Wall-clock (s)} & \textbf{Decode (s)} & \textbf{Forward (s)} & \textbf{GPU mem (GB)} & \textbf{vs.\ 128f} \\
\midrule
text-only & 38 & 0.041 & 0.000 & 0.033 & 15.5 & $-98.2\%$ \\
$16f$ & 66 & 0.542 & 0.088 & 0.417 & 16.0 & $-76.5\%$ \\
$32f$ & 70 & 0.951 & 0.120 & 0.744 & 16.5 & $-58.8\%$ \\
$64f$ & 73 & 1.795 & 0.204 & 1.426 & 17.5 & $-22.2\%$ \\
$\mathbf{128f}$ & $\mathbf{74}$ & $\mathbf{2.308}$ & $0.205$ & $1.901$ & $18.1$ & $0\%$ \\
$256f$ & $74$ & $2.273$ & $0.195$ & $1.874$ & $18.1$ & $-1.5\%$ \\
\midrule
\multicolumn{7}{l}{\textit{Cascade} $16f \to 32f \to 128f$, $\theta = 0.91$} \\
\quad Shared-frame mode & $76.5^{\dagger}$ & $\mathbf{1.658}$ & --- & --- & $17.3$ & $\mathbf{-28.2\%}$ \\
\quad No-share mode    & $76.5^{\dagger}$ & $2.624$ & --- & --- & $17.3$ & $+13.7\%$ \\
\bottomrule
\end{tabular}%
}\\[2pt]
{\footnotesize $^{\dagger}$Accuracy on the $100$-item subset; both cascade modes produce identical predictions by construction. Full-set cascade accuracy is $76.5\%$ (Table~\ref{tab:main}).}
\end{table}

Three observations warrant emphasis:

\paragraph{(i) The marginal wall-clock cost of $128f\!\to\!256f$ is approximately zero.}
The forward time differs by $0.03$s and the GPU memory is identical to within noise, yet accuracy drops from $76.5\%$ (full set) at $128f$ to $75.8\%$ at $256f$.
Under Qwen2.5-VL-7B's video token compression, increasing the frame budget from $128$ to $256$ carries negligible runtime cost---but yields a worse model.
This is not a classic compute--accuracy trade-off but an inference-time regression at essentially zero marginal cost.
Practitioners should prefer $128f$ over $256f$ on this model for accuracy reasons, not efficiency.

\paragraph{(ii) Wall-clock and equivalent-frame savings agree closely.}
The cascade's $31.7\%$ equivalent-frame reduction translates to $28.2\%$ measured wall-clock reduction in shared mode---a gap of only $3.5$ percentage points.
The frame-sharing assumption used in our main-paper cascade accounting is therefore empirically reasonable when the inference infrastructure supports it.

\paragraph{(iii) No-share cascade is a net loss.}
Under the no-share execution model (each stage re-samples and re-encodes its full frame budget), the cascade runs $13.7\%$ \emph{slower} than fixed $128f$.
Stop-distribution analysis reveals the mechanism: $33\%$ of items stop at $16f$ (cheap), $4\%$ stop at $32f$, and $63\%$ escalate through all three stages, paying $16 + 32 + 128 = 176$ equivalent frames of compute.
Under this setting, the $63\%$ bulk of items that reach the final stage amortize both earlier probe costs plus the full target cost, overwhelming the savings from the $37\%$ early stoppers.
Any practical deployment of frame-budget cascade requires either architectural frame sharing or a KV-cache reuse mechanism across stages; without one, cascade is actively harmful for wall-clock cost.

\section{Per-Configuration Pairwise Confusion Counts}
\label{app:pairwise}

\begin{table}[h]
\centering
\caption{Resolution confusion at fixed $64$ frames (Qwen2.5-VL-7B on short): number of items correct at row resolution but wrong at column resolution.}
\label{tab:res_confusion}
\resizebox{0.35\textwidth}{!}{%
\begin{tabular}{lcccc}
\toprule
& \textbf{76K} & \textbf{151K} & \textbf{235K} & \textbf{360K} \\
\midrule
\textbf{76K} & --- & 29 & 27 & 29 \\
\textbf{151K} & --- & --- & 26 & 23 \\
\textbf{235K} & --- & --- & --- & 13 \\
\bottomrule
\end{tabular}%
}
\end{table}

\section{Iso-Budget Pairwise Tradeoffs}
\label{app:isobudget}

At a matched $\sim$9.7M total pixel budget on Qwen2.5-VL-7B short:
\begin{itemize}[nosep,leftmargin=*]
  \item $128f{\times}76K$ correct but $32f{\times}302K$ wrong: $71$ items
  \item $32f{\times}302K$ correct but $128f{\times}76K$ wrong: $38$ items
  \item $128f{\times}76K$ correct but $64f{\times}151K$ wrong: $51$ items
  \item $64f{\times}151K$ correct but $128f{\times}76K$ wrong: $30$ items
  \item $64f{\times}151K$ correct but $32f{\times}302K$ wrong: $52$ items
  \item $32f{\times}302K$ correct but $64f{\times}151K$ wrong: $40$ items
\end{itemize}

These large pairwise tradeoff counts confirm that the temporal-vs-spatial preference is genuinely item-dependent and not merely an artifact of one allocation being uniformly better.

\section{Calibration Details}
\label{app:calibration}

\begin{table}[h]
\centering
\caption{Calibration metrics on Video-MME (Qwen2.5-VL-7B). ECE computed with $10$ bins over the margin-based confidence $\max_k p_k$.}
\label{tab:calibration}
\resizebox{0.7\textwidth}{!}{%
\begin{tabular}{lcccc}
\toprule
\textbf{Split} & \textbf{Config} & \textbf{Accuracy (\%)} & \textbf{Avg.\ Confidence} & \textbf{ECE} \\
\midrule
short & text-only & 37.5 & 0.558 & 0.174 \\
short & $16f$ & 65.9 & 0.712 & 0.087 \\
short & $64f$ & 72.7 & 0.780 & 0.066 \\
short & $128f$ & 76.5 & 0.803 & 0.048 \\
\midrule
medium & text-only & 38.5 & 0.568 & 0.183 \\
medium & $16f$ & 52.3 & 0.651 & 0.128 \\
medium & $64f$ & 64.2 & 0.698 & \textbf{0.068} \\
medium & $128f$ & 66.0 & 0.725 & 0.071 \\
\bottomrule
\end{tabular}%
}
\end{table}

Two observations:
(i)~Short calibration improves monotonically with frame count, matching accuracy gains.
(ii)~Medium calibration is \emph{minimized} at $64f$, not $128f$: while accuracy still rises from $64f$ to $128f$, calibration slightly degrades.
For visually-confused items on short, the mean confidence at the $128f$ wrong answer is $0.55$ (vs.\ a chance-level baseline of $0.25$ for $4$-way MCQA), placing them in a ``moderately confident wrong'' regime that is particularly adversarial to margin-based confidence routing.

\section{Mechanism Hypotheses and Benchmark-Design Implications}
\label{app:mechanism_discussion}

This appendix expands the mechanism boundaries and benchmark-design implications summarized
in \S\ref{sec:mechanisms}.

\subsection{Why Visual Confusion? Multiple Testable Pathways}

The combined evidence does not support a single mechanism. Sampling-sensitive evidence
composition is directly implicated for a subset because changing only the sampling policy
changes correctness and recovers 29.0\% of terminal regressions. Persistent counting cases
are consistent with temporal accumulation or occurrence tracking; salient OCR and dominant-
scene cases motivate competition between an incomplete cue and the queried event or referent.
Attention dilution and distractor competition remain plausible architectural hypotheses, but
the present data do not measure attention causally. The option-length association from the
original univariate analysis (median 9.75 vs. 14.0 characters, $p=4.8\times10^{-4}$,
Cliff's $\delta=-0.19$) also does not survive as an independent causal explanation in the
multivariate permutation audit. Appendix~\ref{app:mechanism_audit} gives the intervention and
case-level evidence.

\subsection{Implications for Benchmark Design}

Three concrete practices follow from our findings.
First, aggregate accuracy should be complemented with per-item correctness traces to expose hidden dynamics (the artifact we release supports this directly).
Second, comparisons across models should use matched configuration grids: on V1 short, Qwen2.5 changes from $19.2\%$ confusion over its 11-configuration surface to $15.9\%$ on the shared four-point grid, while Qwen3 changes from $16.7\%$ over six configurations to $16.3\%$. Unnormalized comparisons conflate scaling ceilings with scaling behavior.
Third, benchmark design should reduce exploitable priors---the $4$-option V1 protocol masks a systematic letter-prior bias that the $8$-option V2 protocol exposes (LLaVA-NeXT-Video-7B picks A--D on $92$--$94\%$ of V2 items, effectively collapsing to a $4$-option prior; Appendix~\ref{app:llava_nextvideo})---and option-coverage diagnostics should be reported alongside aggregate accuracy.

\section{Broader Impact}
\label{app:broader_impact}

\paragraph{Intended positive impact.}
By releasing per-item correctness labels and the cached evaluation pipeline, this work supports more transparent and reproducible evaluation of Video LLMs. Item-level analysis exposes failure modes that aggregate accuracy hides, enabling: (i)~more honest model comparison through matched-grid normalization; (ii)~development of item-aware adaptive inference that can reduce the carbon and compute cost of Video-LLM deployment; and (iii)~better-targeted benchmark design that is robust to letter-prior bias and text-overwrite shortcuts.

\paragraph{Potential risks and misuse.}
We identify three concerns. First, per-item correctness labels can in principle be used to fine-tune or prompt-tune models in a way that overfits to the specific Video-MME / MLVU items in our release; we mitigate this by releasing only model outputs, not the underlying videos, and by providing a matched-grid normalization protocol that resists such overfitting. Second, our visual-confusion analysis quantifies failure modes that could be exploited adversarially (e.g., constructing inputs that flip Video-LLM answers); however, the failure modes we document are intrinsic to current architectures rather than novel attack vectors. Third, the cascade reference baseline reduces compute but cannot eliminate the item-level oracle gap; deployers should not interpret cascade savings as a justification for replacing strong-but-expensive models when stakes are high.

\paragraph{Energy and compute considerations.}
Our experiments evaluate five open Video LLMs across MCQA and generation settings, with
approximately 0.13M released per-item records. Reusing frame caches avoids redundant decoding
for cached coverage cells, while raw-video cells are retained when required for protocol
replication. We report measured wall-clock and memory for the deployment reference in
Appendix~\ref{app:runtime}; we do not extrapolate those measurements to a project-wide energy
percentage.

\paragraph{Data and license boundaries.}
We do not redistribute underlying benchmark media or question text. Code is released under
MIT and newly created model outputs and annotations under CC-BY-4.0; benchmark-derived fields
remain subject to their source terms. The artifact contains no newly collected human-
participant data or person-descriptive attributes.

\paragraph{What we explicitly do not enable.}
The released artifact does \emph{not} include: (i)~surveillance-relevant model fine-tuning recipes; (ii)~training data for facial-recognition or person-identification systems; (iii)~harmful content classifiers. Our analysis target is a narrow technical question about scaling behavior, not deployment-readiness for any downstream user-facing application.

\section{Source Governance and Derived-Artifact Boundary}
\label{app:ethics}

Table~\ref{tab:source_terms} records the source-provider terms checked for this release. This
is a provenance summary rather than a relicensing claim. The derived-artifact license applies
only to components created by this study and does not supersede benchmark or media-owner
rights.

\begin{table}[h]
\centering
\caption{Source governance and our release boundary. Terms are summarized from the linked
provider pages; users must consult the current source text.}
\label{tab:source_terms}
\resizebox{\linewidth}{!}{%
\begin{tabular}{p{0.16\linewidth}p{0.39\linewidth}p{0.37\linewidth}}
\toprule
\textbf{Source} & \textbf{Provider-stated terms} & \textbf{Handling in this artifact} \\
\midrule
Video-MME v1 & Academic research only; commercial use prohibited; video copyrights remain
with owners; redistribution requires prior approval. & No videos, frames, subtitles, audio,
or question text. Derived rows retain source provenance. \\
Video-MME v2 & Academic research only; commercial use prohibited; video copyrights remain
with owners; redistribution requires prior approval. & Same boundary as v1; v2 labels are
protocol-tagged separately. \\
MLVU & CC-BY-NC-SA-4.0 and research-only notice; benchmark authors state that they do not own
raw-video copyrights and provide a removal channel. & No source media or question text;
generation and MCQA records remain source-tagged. \\
AVSD / Charades & AVSD annotations and Charades videos are obtained through their official
provider channels and remain governed by those providers' terms. & No source videos, audio,
dialogue text, or reference answers are redistributed. \\
\bottomrule
\end{tabular}%
}
\end{table}

Provider pages are
\url{https://github.com/MME-Benchmarks/Video-MME},
\url{https://github.com/MME-Benchmarks/Video-MME-v2},
\url{https://github.com/JUNJIE99/MLVU}, and the AVSD project associated with
\citet{alamri2019audio}. If a source provider withdraws an item, the maintenance policy is to
remove or tombstone the corresponding derived identifier. Some source videos may depict
identifiable people; responsibility for original collection and media rights remains with the
source providers, while this work is responsible for minimizing exposure in its derived
release. The artifact's task and trajectory tags describe model behavior and do not encode
demographic or person-identifying attributes.

\end{document}